\documentclass{article}

\usepackage{todonotes}

 \usepackage[main, final]{neurips_2026}
\usepackage[utf8]{inputenc} 
\usepackage{hyperref}       
\usepackage[T1]{fontenc}    
\usepackage{url}           
\usepackage{booktabs}       
\usepackage{amsfonts}       
\usepackage{nicefrac}       
\usepackage{microtype}      
\usepackage{xcolor}         

\usepackage{multicol}
\usepackage{wrapfig}

\usepackage{times}
\usepackage{soul}
\usepackage[small]{caption}
\usepackage{graphicx}
\usepackage{amsmath}
\usepackage{amsthm}
\usepackage{algorithm}
\usepackage{algorithmic}
\usepackage{bm}
\usepackage{multirow}
\usepackage{graphbox}
\usepackage{subcaption}

\usepackage{amsmath,amsfonts,bm}

\def\eqref#1{equation~\ref{#1}}

\def\1{\bm{1}}

\def\rvn{{\mathbf{n}}}

\def\rvx{{\mathbf{x}}}
\def\rvy{{\mathbf{y}}}

\def\vf{{\bm{f}}}

\def\vk{{\bm{k}}}

\def\vx{{\bm{x}}}
\def\vy{{\bm{y}}}

\def\mC{{\bm{C}}}
\def\mD{{\bm{D}}}
\def\mE{{\bm{E}}}
\def\mF{{\bm{F}}}

\def\mM{{\bm{M}}}

\DeclareMathAlphabet{\mathsfit}{\encodingdefault}{\sfdefault}{m}{sl}
\SetMathAlphabet{\mathsfit}{bold}{\encodingdefault}{\sfdefault}{bx}{n}
\newcommand{\tens}[1]{\bm{\mathsfit{#1}}}

\def\tD{{\tens{D}}}

\def\tF{{\tens{F}}}

\def\tN{{\tens{N}}}

\def\tX{{\tens{X}}}
\def\tY{{\tens{Y}}}

\newcommand{\E}{\mathbb{E}}

\newcommand{\R}{\mathbb{R}}

\newcommand{\Var}{\mathrm{Var}}

\DeclareMathOperator*{\argmin}{arg\,min}

\title{Filter-Enhanced Smartphone Hyperspectral Imaging}

\author{%
  Daniil Reutsky\textnormal{\textsuperscript{ 1}}
  \And
  Daniil Vladimirov\textnormal{\textsuperscript{ 2, 3, 4}}
  \And
  Yasin Mamedov\textnormal{\textsuperscript{ 2, 4}}
  \And
  Georgy Perevozchikov\textnormal{\textsuperscript{ 1}}
  \And
  Egor Ershov\textnormal{\textsuperscript{ 2, 3, 4}}
  \And
  Nancy Mehta\textnormal{\textsuperscript{ 1}}
  \And
  Radu Timofte\textnormal{\textsuperscript{ 1}}
}

\begin{document}
\maketitle
\vspace{-2em}
\begin{center}
    \mbox{\textsuperscript{1} Computer Vision Lab, CAIDAS \& IfI, University of W\"urzburg, Germany}\quad
    \mbox{\textsuperscript{2} MIRAI, Moscow, Russia}\quad
    \mbox{\textsuperscript{3} AXXX, Moscow, Russia}\quad
    \mbox{\textsuperscript{4} Color Reproduction and Synthesis Institute, Moscow, Russia}
\end{center}
\vspace{2em}
\begin{abstract}
Hyperspectral reconstruction from RGB imagery offers an affordable alternative to costly hyperspectral acquisition devices.
However, its accuracy is fundamentally constrained by the limited spectral resolution of standard RGB cameras.
In this paper, we propose a new approach that leverages the multi-camera systems integrated into most modern smartphones and enhances their capabilities by modulating their spectral response functions using carefully selected spectral filters, enabling a smartphone to capture richer spectral information.
We propose a neural network able to exploit auxiliary RGB images with diverse spectral characteristics, enabling more accurate hyperspectral reconstruction compared to conventional single-image methods.
Training is supported by a newly collected dataset, Doomer, which consists of multiple misaligned RGB images alongside corresponding hyperspectral data.
We also propose a device-adaptation strategy that conditions the multi-image hyperspectral reconstruction network on camera color correction matrices available in standard DNG metadata, requiring no real images from the target device.
We evaluate it on three physical smartphones, two of them unseen during training.
Code is available at \url{https://github.com/cromtus/filters}
\end{abstract}

\begin{figure*}[ht]
    \centering
    \includegraphics[width=0.8\textwidth]{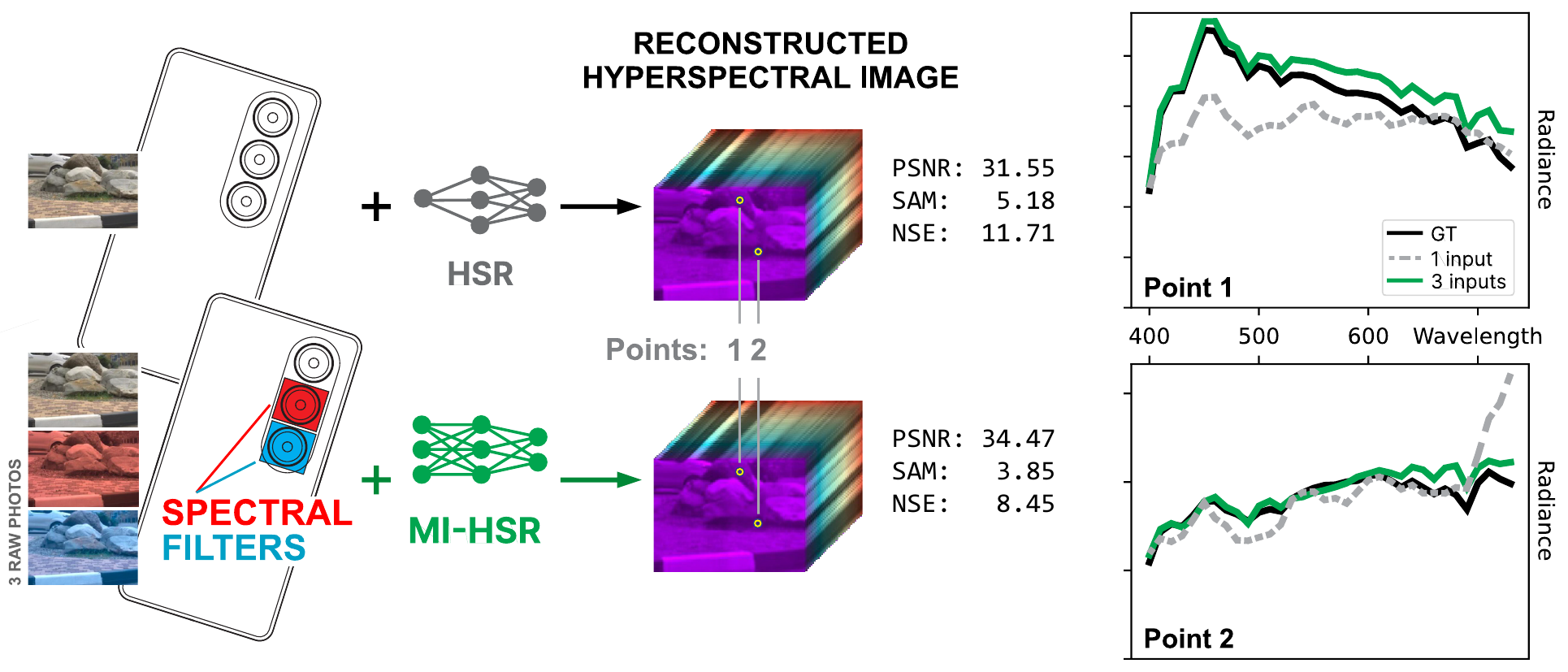}
    \caption{
    Proposed low-cost mobile spectral imaging system that transforms a standard smartphone into a spectrally diverse capture device via external filters on auxiliary cameras. This configuration enables 
    multi-channel acquisition without internal hardware modification, supporting practical and scalable hyperspectral reconstruction.}
    \label{fig:abstract}
\end{figure*}
    
\section{Introduction}
Hyperspectral imaging provides dense spectral measurements at each spatial pixel, forming a 3D cube $\tY \in \mathbb{R}^{h\times w\times n}$, where $n \gg 3$.
This enables fine-grained analysis of material properties in applications ranging from remote sensing~\citep{remote-sensing1}, to medical diagnostics~\citep{medical-imaging}, to historical preservation~\citep{documents}, to ISP improvement~\citep{mobile-spec, beyond-rgb}, to food quality assessment~\citep{food-review, food-tomato}.
However, acquiring such high-dimensional data typically requires hardware that is expensive, bulky, and often reliant on time-consuming scanning thus fundamentally limiting the usability of hyperspectral imaging in dynamic or consumer settings.

An increasingly studied alternative is \emph{hyperspectral reconstruction} (HSR): recovering $\tY$ from an RGB observation $\tX \in \mathbb{R}^{h \times w \times 3}$ captured via a sensor with a spectral sensitivity matrix $\mC \in \mathbb{R}^{n \times 3}$ under noise realization $\tN \in \mathbb{R}^{h \times w \times 3}$:
\[
\tX =  \tY\mC + \tN.
\]
This inverse problem is highly ill-posed and despite advances in deep learning-based HSR~\citep{mst++, hrnet}, reconstructing $\tY$ from a single RGB view remains fundamentally limited by the low spectral observability.

Several efforts have aimed to improve this observability. Learned or optimized multispectral filter arrays (MSFAs)~\citep{msfa-selection}, end-to-end spectral sensitivity learning~\citep{dl-filter}, and joint sensor-network co-design~\citep{joint} have all been proposed.
However, these methods assume control over sensor hardware and manufacturing, thus making them impractical for scalable or consumer-level deployment. 

One underexplored but promising path is to leverage the multi-camera systems found in modern smartphones.
In principle, such devices can be treated as a low-cost, multi-spectral capture system thus capable of observing a scene through multiple, distinct sensitivity matrices. 
However, prior works on leveraging multiple cameras~\citep{shrestha2010multispectral, diy-brown, mobispectral} do not tackle unavoidable image misalignment.

\emph{Can we turn a multi-camera smartphone into a compact, spectrally diverse imaging system, without altering its internal hardware?}
Our key insight is that by modulating the spectral response of the auxiliary cameras using carefully chosen external filters in combination with a new transformer-based neural network, we can create an imaging system that allows richer spectral representations.
Unlike MSFAs or custom hardware, our approach requires no internal modification and potentially enables real-time, parallel acquisition via standard multi-camera APIs~\citep{apple2024multicam}.
Critically, this setup is easy to deploy, manufacturable at scale, and compatible with existing mobile infrastructure.

The resulting setup produces spatially misaligned but spectrally rich multi-view data thus posing a new fusion problem that we address via alignment-aware learning.
{An overview of our physical configuration is shown in Fig.~\ref{fig:abstract}}, which illustrates how external filters are applied to the auxiliary lenses of a standard smartphone to create a spectrally diverse input set.
The combination of low-cost physical augmentation and learning-based reconstruction represents a practical path toward deployable hyperspectral imaging in unconstrained environments.

We propose a complete, low-cost approach for \emph{multi-image-to-hyperspectral reconstruction (MI-HSR)} using a filter-enhanced smartphone.
Specifically, the contributions are three-fold:

\begin{itemize}
    \item A novel, low-cost spectral acquisition approach that converts a commodity triple-camera smartphone into a hyperspectral imaging device. We also demonstrate how this approach can be utilized for other triple-camera devices using standard DNG metadata stored with every RAW image.

    \item A multi-image hyperspectral reconstruction model that combines established components, including optical-flow-guided deformable convolutions and a gated feed-forward network, to align features from misaligned auxiliary views. The proposed reconstruction model, together with the novel acquisition approach, achieves a 2.92~dB PSNR improvement over the strongest single-image method.

    \item The Doomer dataset, the first real-world benchmark for the MI-HSR problem.
    The dataset contains 155 real-world scenes, each providing spatially misaligned RAW images from the filtered smartphone setup and a corresponding ground-truth hyperspectral scan.

\end{itemize}

\section{Related Work}

\textbf{Low-cost multispectral imaging}\quad
Numerous approaches have aimed to capture multispectral or hyperspectral information without expensive hardware.
Early work by \citet{wheel7} employed a grayscale camera with a rotating filter wheel, while \citet{valero2007} used an RGB camera with three interchangeable filters.
\citet{diy-brown} captured scenes with three different cameras, leveraging the variation in their spectral sensitivities.
More recently, \citet{mobispectral} demonstrated that consumer mobile devices with both RGB and NIR sensors can achieve extended spectral capture (400--1000~nm).
Although these systems reduce costs, they either involve long capture times or assume no misalignment between successive captures.

\textbf{Hyperspectral reconstruction from RGB}\quad
Traditional HSR methods model spectra using sparse coding~\citep{arad-icvl}, dictionary learning~\citep{a-plus}, or manifold embeddings~\citep{manifold-mapping}, based on the low-dimensional structure of hyperspectral data and the rarity of metamers~\citep{foster-metamers}.
These approaches are computationally efficient but often lack the capacity to incorporate global context from the input image, making them less robust to complex natural scenes.
Recent methods based on deep learning have achieved significant advances~\citep{ntire-2022}.
Early deep models used CNNs~\citep{galliani, hscnn-plus}, while newer transformer-based approaches like MST++~\citep{mst++} and SPECAT~\citep{specat} introduced attention along spectral or spatial dimensions, and diffusion-based approaches like SDM \citep{nips-metamers} or LDERT \citep{ldert} have also been adapted for HSR.
However, nearly all research relies on synthetic RGB inputs rendered from hyperspectral images (HSIs) using color matching functions (CMFs), assuming perfect alignment and sometimes access to camera parameters.
These assumptions do not hold in practice, limiting model generalizability and making it non-sustainable to hard metameric cases \citep{fu-limitations, nips-metamers}.
We address these limitations by training and evaluating on real-world data with acquisition artifacts and misalignments.

\textbf{Hyperspectral datasets}\quad
Early datasets like CAVE~\citep{cave-dataset} and Harvard~\citep{harvard-dataset} provided controlled hyperspectral measurements.
Later datasets such as ICVL~\citep{arad-icvl}, KAUST~\citep{kaust-dataset}, and ARAD~1K~\citep{ntire-2022} focused on enabling data-driven HSR methods.
These datasets contain either radiance or reflectance data, but most lack real RGB images and instead simulate RGB via given sensor spectral sensitivities, which fails to capture the characteristics of camera pipelines.
Moreover, they often assume perfect alignment, which does not hold in practical settings.
Recent datasets like BeyondRGB~\citep{beyond-rgb} and MobileSpec~\citep{mobile-spec} address some of these issues by including real RGB captures.
However, they still face trade-offs between alignment, diversity, and color reference availability, see comparison Table~\ref{tab:datasets} in Supp. 
We build on these efforts by introducing the Doomer dataset with real misaligned RGB images, hyperspectral data, and in-scene color references thus enabling realistic reconstruction under natural capture conditions.

\textbf{Handling misalignment}\quad
Misalignment between inputs is a well-known challenge in video and reference-based super-resolution tasks.
Optical flow \citep{sttn, basicvsr}, deformable convolutions~\citep{tdan, edvr}, attention mechanisms~\citep{dual} have been proposed to mitigate this.
In our context, misalignment also arises due to ground truth HSIs being not spatially aligned with the RGB input.
\citet{raw-to-srgb} addressed this by warping ground truth toward the input using optical flow, enabling pixel-level evaluation.
\citet{mimicked} proposed contextual losses and pseudo-aligned inputs as training strategies.
We adopt the interpretable and evaluation-friendly approach~\citep{raw-to-srgb} of warping the ground truth to the input using optical flow, allowing accurate pixel-wise supervision and computation of metrics.
\section{Proposed Imaging System}
\label{sec:proposed}
\subsection{System Overview}


To amplify spectral diversity, we cover Tele and Wide cameras with external filters.
As illustrated in Fig.~\ref{fig:abstract}, this converts each RGB camera into a spectrally modulated sensor.
The resulting device captures nine distinct spectral channels without requiring scanning or hardware changes.

Formally, let $\mC_i \in \mathbb{R}^{n \times 3}$ be the spectral sensitivity functions (SSF) of camera $i$, and $\bm{f}_i \in [0,1]^n$ the spectral transmittance of the filter applied to that camera.
Thus, per-camera response becomes $\mC_i \odot \vf_i$, and the overall response is:
\[
    \mC_{\mF} =
    \begin{bmatrix}
        (\mC_0 \odot \vf_0);
        \hdots;
        (\mC_{k-1} \odot \vf_{k-1})
    \end{bmatrix} \in \mathbb{R}^{n \times 3k},
\]
where $\mF = [\vf_0; \dots; \vf_{k-1}]$ is the filter configuration and $\odot$ is element-wise multiplication.
In our prototype, $k = 3$, with one unfiltered camera ($\vf_0 = (1, \dots, 1)^\top$) and two filtered cameras.
This design simplifies dataset collection (Sec.~\ref{sec:dataset}) while still significantly enriching spectral representations.

This configuration has two practical advantages.
First, all channels could be captured simultaneously under natural illumination, making it suitable for dynamic scenes.
Second, it relies entirely on off-the-shelf hardware components.
However, as each camera has a distinct physical position, the resulting images are spatially misaligned.
This necessitates learning-based alignment modules, which we incorporate into our reconstruction pipeline (Sec.~\ref{sec:method}).

\subsection{Filter Selection via Spectral Uncertainty Minimization}
\label{sec:filter-selection}

The effectiveness of our imaging system critically depends on the choice of spectral filters.
Since we train on fully real-world data, the filter configuration must be fixed prior to data collection.

Given a library of $65$ candidate filters from a standard set, we exhaustively evaluate all $65 \times 64$ ordered filter pairs for the two auxiliary cameras.
The optimal pair should minimize spectral ambiguity --- i.e., the uncertainty in the latent spectrum $\rvy \in \R^n$ given a measurement $\rvx = \mC_{\mF}^\top \rvy + \rvn$, where $\rvn$ is sensor noise.

Following~\cite{reutskii2024spectral}, we use the expected conditional variance of the spectrum as a selection criterion:
\[
    v(\mF) = \E_{\rvx} \left[ \mathrm{tr} \, \Var_{\rvy}(\rvy \mid \rvx) \right].
\]
This criterion reflects the average spectral uncertainty remaining after observing $\rvx$.
Lower $v(\mF)$ implies more informative measurements and improved reconstructability.

To compute this metric, we sample spectra $\mathbf{y}$ from a uniform distribution over pixels in a publicly available hyperspectral dataset~\citep{kaust-dataset}.
We also sample SSFs via a random convex combination of 7 known SSFs with respect to spectrophotometer-measured filter transmittances (see Sec.~\ref{sec:supp:measurements} for details).
The final filter pair (Fig.~\ref{fig:best-filters} in Supp.) is selected as the one that minimizes $v(\mF)$.
By not adhering to specific SSFs, we ensure that the proposed filter pair is device-agnostic compromise rather than optimum for any specific smartphone.

To verify effectiveness of the proposed filter selection strategy, we trained a MI-HSR method in simulated settings with 12 different filter sets.
See Sec.~\ref{sec:supp:filters} for details.
\section{Doomer Dataset}
\label{sec:dataset}

\paragraph{Overview} To support training and evaluation for the MI-HSR task introduced in Sec.~\ref{sec:proposed}, we collect a new dataset, which we call \textit{Doomer}\footnote{The name \textit{Doomer} is inspired by the subcultural aesthetic: most scenes were collected under gloomy or overcast weather conditions, in contrast to the brightly lit existing datasets.}.
Existing hyperspectral datasets are not suited for our setup: none provide spatially misaligned multi-view real RGB observations.

Doomer contains 155 real-world scenes captured using a Huawei Mate 40 Pro smartphone equipped with Main, Tele, and Wide cameras, along with a Specim IQ hyperspectral camera for ground truth.
In each scene, we record three smartphone RAW images, two with custom spectral filters and one unfiltered as well as a 111-band (400--730 nm) hyperspectral image.

To provide representability, scenes include a mix of indoor and outdoor environments under varying illumination conditions (e.g., halogen and LED lighting indoors; overcast and sunny weather outdoors), and span a range of objects including food, printed material, architectural surfaces, and color calibration charts.
Each scene includes a gray ball reference for future work on illumination estimation.
Gray ball reference makes the proposed dataset potentially useful in research on illumination estimation, automatic white balance and color space transform (see Sec.~\ref{sec:supp:dataset} for more detailed proposals).

\paragraph{Capture setup} The data acquisition rig consists of a Specim IQ hyperspectral camera, a Huawei Mate 40 Pro smartphone mounted in a 3D-printed case with slots for spectral filters and a gray reference sphere (VFX ball).

\begin{figure}[ht]
    \centering
    \includegraphics[width=0.8\linewidth]{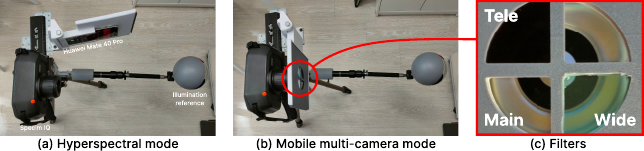}
    \caption{\textit{Capture setup for the Doomer dataset}. (a)~Smartphone holder rotated to allow hyperspectral capture via Specim IQ. (b)~Smartphone repositioned for 
    multi-camera RGB capture. (c)~External spectral filters mounted on Tele and Wide cameras to induce spectral diversity.}
    \label{fig:setup}
\end{figure}

The entire system is mounted on a tripod, with the phone positioned to rotate along a vertical axis for alternating captures (Fig.~\ref{fig:setup}).
This design allows the smartphone and hyperspectral camera to image scenes from nearly identical viewpoints, minimizing parallax and occlusion.

Image acquisition proceeds sequentially: first, all three smartphone cameras capture RAW images; then, the smartphone case is moved to capture hyperspectral image.
Most smartphone settings (e.g., ISO, shutter speed) are controlled automatically, except in scenes with poor red signal where we manually adjusted Tele exposure to avoid excessive noise.

\paragraph{Data preprocessing} Each four-image scene group (three RGB + one HSI) undergoes standardized preprocessing (see Sec.~\ref{sec:supp:dataset}).
We also normalize field of view and resolution across sensors.
Specifically, we estimate pairwise homographies between each RGB image and the Tele view using SIFT keypoints and RANSAC.
All images are then warped to the Tele frame and cropped accordingly.
When automatic registration fails, manual alignment is used.
Despite this correction, residual geometric misalignment remains due to parallax and non-planar scene structure --- motivating our use of alignment-aware HSR models.
Finally, all images are downsampled to match the hyperspectral resolution, originally $194 \times 259$ cropped to $192 \times 256$ for convenience, and spectral grid is resampled to conventional 400--730 nm, $n=34$, to optimize computational costs.
\section{Method}
\label{sec:method}

The task of MI-HSR involves predicting a hyperspectral image $\tY \in \mathbb{R}^{h\times w\times n}$ aligned to a reference RGB image $\tX_0 \in \mathbb{R}^{h\times w\times 3}$ given also auxiliary RGB images $\{\tX_i\}_{i=1}^{k-1} \subset \mathbb{R}^{h\times w\times 3}$ from multiple spatially offset sensors.
This setup introduces two key challenges: (i) the available hyperspectral supervision $\tY$ is not aligned with any RGB input, and (ii) the input views are misaligned due to differing camera geometries.
Our approach addresses both issues, one through pre-processing, the other through architectural design.

\subsection{Supervision Warping via Optical Flow}
\label{sec:method:supervision}

To leverage the target hyperspectral image for training, we align it to the reference RGB view using pre-trained optical flow (OF).
Since spectral and RGB images differ in modality, we first compute a color projection:
\[
\mC_{\mathrm{Y}\to0} := \argmin_{\mC \in \mathbb{R}^{n \times 3}} \|\tY \mC - \tX_0\|_2^2 + \mu\|\mC\|_2^2.
\]
$\tY \mC_{\mathrm{Y}\to0}$ serves as an approximation of $\tY$ in the color space of the reference RGB.
We noted that the approximation becomes noisy in the absence of the regularization term due to ill-posedness of the problem.
Since now both views share the same color space, we can estimate a dense correspondence field $\tD_{0\to\mathrm{Y}} \in \mathbb{R}^{h \times w \times 2}$ using a pre-trained optical flow model~\citep{pwc}:
\[
\tD_{0\to\mathrm{Y}} := \mathrm{OpticalFlow}(\tX_0, \tY \mC_{\mathrm{Y}\to0}).
\]
This flow is used to warp the hyperspectral ground truth (GT) to the reference view:
\begin{equation*}
\tY_\mathrm{w} := \mathrm{Warp}(\tY, \tD_{0\to\mathrm{Y}}), \quad \mM := \lfloor \mathrm{Warp}(\mE, \tD_{0\to\mathrm{Y}}) \rfloor,
\end{equation*}
where $\mM$ is a binary mask indicating valid visible pixels, $\mE$ is a matrix of ones.
This enables aligned supervision for training and point-wise loss computation $\mathcal{L}_1(\hat{\tY} \odot \mM, \tY_\mathrm{w} \odot \mM)$

\subsection{Deformable Convolution Alignment}

Even with warped supervision, the three input views remain spatially misaligned.
Direct flow-based registration is possible but leads to unnecessary error accumulation.
Instead, one would prefer leveraging the misaligned views ``as-is'' in some flow-aware manner.
To this end, we make use of deformable convolutions~\citep{dconv}, which we feed with a dense correspondence field between inputs computed by the optical flow model.
This allows the convolution layer to deterministically shift each of its 3×3 sampling positions toward geometrically matched locations at an auxiliary view and produce a feature map aligned to the reference view.
The kernel weights remain learnable, which allows the filter to handle residual local misalignments and simultaneously learn features useful for further spectral reconstruction.

Similar to the previous section, we first unify color spaces of the reference and auxiliary inputs and compute optical flow between them:

\begin{align*}
\mC_{0\to i} &:= \argmin_{\mC \in \mathbb{R}^{3 \times 3}} \|\tX_0 \mC - \tX_i\|_2^2,\quad & i = 1, ..., k-1,\\
\tD_{0\to i} &:= \mathrm{OpticalFlow}(\tX_0\mC_{0\to i}, \tX_i),\quad & i = 1, ..., k-1.
\end{align*}

This time we do not need regularization because channels in $\tX_0$ are not as highly correlated as in $\tY$.
The reference input $\tX_0$ is then passed through a standard 3×3 convolution layer, while the auxiliary inputs $\tX_i$ are passed through the aforementioned deformable convolution together with their corresponding deformation fields:

\[
\tF_0 := \mathrm{Conv}(\tX_0),\quad
\tF_i := \mathrm{DeformConv}(\tX_i, \tD_{0\to i}).
\]

Because $\mathrm{DeformConv}$ employs a 3×3 kernel, it needs 3×3 sampling points per output pixel.
The deformation field supplies only the central offset; the other eight offsets are fixed relative shifts (-1,-1), (-1,0), ..., (+1,+1) applied around it.
Concatenation of all feature maps $\{\tF_i\}_{i=0}^{k-1}$ serves as the input for the Spectral Reconstruction module.

We use a pre-trained PWC-Net~\citep{pwc} with frozen weights as the main optical flow model in DCAM.
An ablation study on the choice of the optical flow model is presented in Sec.~\ref{sec:supp:of-ablation}.

\subsection{Spectral Reconstruction}

We employ a UNet-like architecture Gated Multi-Stage Transformer (GMST++) heavily inspired by MST++~\citep{mst++}.
Its main building block is Spectral Transformer (ST) block, which in turn consists of Spectral Multi-head Self-attention (S-MSA) and the Gated DConv Feed-forward Network (GDFN) adopted from Restormer~\citep{restormer}.
S-MSA treats each 2D slice of the feature tensor as a token and computes attention across channels based on their global correlations.
GDFN comprises two branches of successive 1×1 and depth-wise 3×3 convolutions; one branch is GeLU-activated and then multiplied by the other, thereby attenuating some spatial locations and amplifying others.
This is particularly helpful in multi-image setting, where small residual misalignments may remain, and their attenuation prevents them from misleading the reconstruction process.

\begin{figure}[h]
    \centering
    \includegraphics[width=0.8\linewidth]{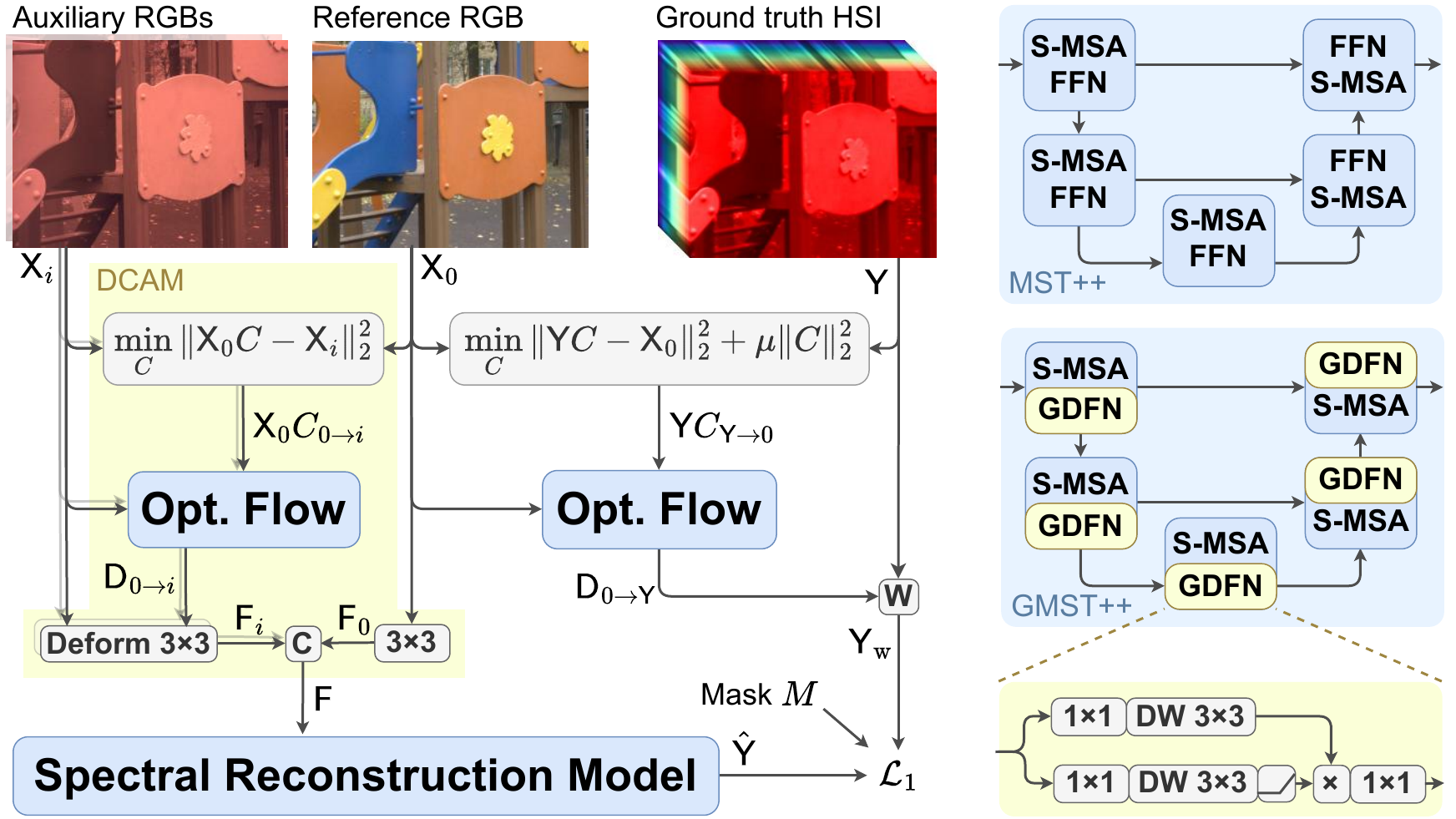}
  \caption{(Left) Our proposed framework with the Deformable Convolution Alignment Module (DCAM) for solving MI-HSR; any existing HSR network can be plugged. (Right) The proposed GMST++ model featuring GDFN module~\citep{restormer}.}

    \label{fig:method}
\end{figure}

\section{Experiments}
Our main intention is to demonstrate the advantage of leveraging a multi-camera imaging system for HSR over reconstruction from a single RGB alone.
We evaluate this in both Clean and Real-world settings.
The Clean setting consists of simulating perfectly aligned RGB images from a widely recognized HSI benchmark.
Such a controlled setup allows us to isolate the effect of increasing the number of cameras and serves as a proof of concept for our proposed acquisition system.
The Real-world setting verifies our misalignment-aware framework and shows that the advantages of a multi-camera system can be retained even in an uncontrolled environment.

\subsection{Experimental Setup}
\label{sec:exps-setup}

\paragraph{Datasets} The Clean setting employs ARAD~1K dataset~\citep{ntire-2022} as the source of ground truth HSIs.
We simulate three input RGBs given camera response functions of Samsung Galaxy~S20, Google Pixel, and iPhone~11~\citep{smartphone-sens}.
The spectral filter set was selected specifically for this camera configuration following the methodology described in Sec.~\ref{sec:filter-selection}, including leaving the first camera without the filter.
Sensor noise was injected according to Gaussian-Poisson model~\citep{noise-model}.
The Real-world setting employs our Doomer dataset as it is currently the only available dataset suitable for MI-HSR.
We split Doomer into 108 training, 20 validation, and 27 held-out test scenes and use the author-provided split for ARAD~1K.

\paragraph{Methods} We choose several state-of-the-art HSR methods for comparison with GMST++: AWAN~\citep{awan}, MST++~\citep{mst++}, and SPECAT~\citep{specat}.
Among general image restoration models, we additionally include NAFNet~\citep{nafnet}.
To evaluate these models in Real-world Multi-image setting, we prepend them with our DCAM module.

\paragraph{Metrics} We adopt the well-known Peak Signal-to-Noise Ratio (PSNR) and Spectral Angle Mapper (SAM). We additionally introduce the Normalized Spectral Error (NSE), which serves as an alternative to MRAE since it does not penalize small values of individual bands and better reflects the integral nature of radiance: $\mathrm{NSE}(\hat{\vy}, \vy) = \frac{||\hat{\vy} - \vy||_1}{||\vy||_1}\cdot 100\%,\quad \hat{\vy}, \vy \in \R^n$.
In the warped GT, some pixels become missing or invalid, as indicated by the mask $\mM$. 
Those pixels are excluded from computation of the metrics.

\paragraph{Implementation details} In all the experiments we used Adam optimizer with learning rate of 0.0004.
During training, each image was randomly cropped to 64×64 and optionally flipped vertically or horizontally.
For the Clean setting, the model was trained for 2,000 epochs, and the final model was selected at the end of training.
For the Real-world setting, the model was trained for 10,000 epochs, with an evaluation loop performed every 30 epochs.
The final model was selected based on the lowest mean absolute error (MAE) on the validation set.

\subsection{Results}
\label{sec:eval:real}
We trained and tested all HSR methods using two input configurations: single-image $\{ \tX_0 \}$ and multi-image $\{ \tX_0, \tX_1, \tX_2 \}$.
The quantitative results for the Clean and Real-world settings are presented in Tables~\ref{tab:clean-world} and~\ref{tab:real-world}, respectively.
Note that we report the number of \emph{trainable} parameters which means the pre-trained weights of PWC-Net are not counted.
However, we do count PWC-Net's contribution in GMACs.

\begin{table}[ht]
\centering
\begin{tabular}{@{}lrrrrr@{}}
\toprule
Method & PSNR, dB ↑ & \multicolumn{1}{c}{SAM, ° ↓} & \multicolumn{1}{c}{NSE, \% ↓} & Params, K & GMACs \\
\midrule
\emph{\textcolor{gray}{Single-image input}} \\
AWAN & 33.91 & 4.19 & 7.31 & 436 & 84.82 \\
NAFNet & 33.84 & \underline{4.14} & 7.36 & 376 & 7.74 \\
SPECAT & 33.63 & 4.17 & 7.34 & 351 & 53.32 \\
MST++ & \underline{33.98} & \textbf{4.06} & \textbf{7.12} & 546 & 31.20 \\
GMST++ (ours) & \textbf{33.99} & 4.15 & \underline{7.25} & 546 & 31.30  \\
\midrule
\emph{\textcolor{gray}{Multi-image input}} \\
AWAN     & \underline{41.98}  & 2.69 & 4.76 & 437 & 84.82\\
NAFNet & 41.27 & \underline{2.63} & 4.83 & 376 & 7.74 \\
SPECAT & 39.44 & 3.05 & 5.54 & 354 & 54.22 \\
MST++   & 41.78 & 2.56 & \underline{4.63} & 548 & 31.62 \\
GMST++ (ours) & \textbf{42.17} & \textbf{2.47} & \textbf{4.51} & 547 & 31.71 \\ \bottomrule
\end{tabular}
\vspace{0.2em}
\caption{Results in Clean setting (ARAD~1K dataset). \textbf{Best} and \underline{second-best} results are highlighted.}
\label{tab:clean-world}
\end{table}

\begin{table}[ht]
\centering
\begin{tabular}{@{}lrrrrr@{}}
\toprule
Method & PSNR, dB ↑ & \multicolumn{1}{c}{SAM, ° ↓} & \multicolumn{1}{c}{NSE, \% ↓} & Params, K & GMACs \\
\midrule
\emph{\textcolor{gray}{Single-image input}} \\
AWAN & 30.48 ± 0.70 & 6.41 ± 1.09 & 13.68 ± 1.79 & 525 & 25.82 \\
NAFNet & 30.63 ± 0.66 & 6.00 ± 0.72 & 13.29 ± 0.98 & 376 & 1.50 \\
SPECAT & 31.08 ± 0.86 & 5.47 ± 0.68 & 12.28 ± 1.25 & 394 & 11.21 \\
MST++~ & \underline{31.33 ± 0.62} & \underline{5.24 ± 0.42} & \underline{12.03 ± 0.82} & 655 & 7.35 \\
GMST++ (ours) & \textbf{31.55 ± 0.51} & \textbf{5.18 ± 0.29} & \textbf{11.71 ± 0.73} & 654 & 7.37 \\
\midrule

\emph{\textcolor{gray}{Multi-image input}} \\
DCAM + AWAN & 32.26 ± 0.58 & 4.35 ± 0.28 & 10.67 ± 0.74 & 527 & 44.06 \\
DCAM + NAFNet & 33.17 ± 0.68 & 4.23 ± 0.46 & 9.80 ± 0.90 & 377 & 19.69 \\
DCAM + SPECAT & 33.45 ± 0.72 & 4.05 ± 0.20 & 9.59 ± 0.70 & 397 & 29.54 \\
DCAM + MST++ & \underline{33.82 ± 0.30} & \underline{3.95 ± 0.25} & \underline{9.09 ± 0.44} & 657 & 25.59 \\
DCAM + GMST++ (ours) & \textbf{34.47 ± 0.77} & \textbf{3.85 ± 0.25} & \textbf{8.45 ± 0.82} & 657 & 25.61 \\
\bottomrule
\end{tabular}
\vspace{0.2em}
\caption{Results in Real-world setting (Doomer dataset). 
All methods are supervised by warped GT ($\tY_{\mathrm{w}}$). 
To let the methods handle multiple misaligned inputs, we prepend them with our DCAM module.
For each method, mean and double std across 10 training runs is reported.
\textbf{Best} and \underline{second-best} results are highlighted.
Same method can have different params and GMACs compared with those reported in Table~\ref{tab:clean-world}, because the number of output channels was increased from 31 to 34 and the input resolution was reduced from 482×512 to 192×256.
}
\label{tab:real-world}

\end{table}

Table~\ref{tab:clean-world} shows the advantage of our GMST++ in multi-image configuration within the Clean setting, though it outperforms its nearest competitor only by 0.2 dB.
Overall, multi-image input allows substantially better reconstruction than from single-image suggesting that our filtering device design unveils lots of spectral information previously hidden in single RGB inputs.

\begin{figure*}[!ht]
    \centering
    \includegraphics[width=0.8\textwidth]{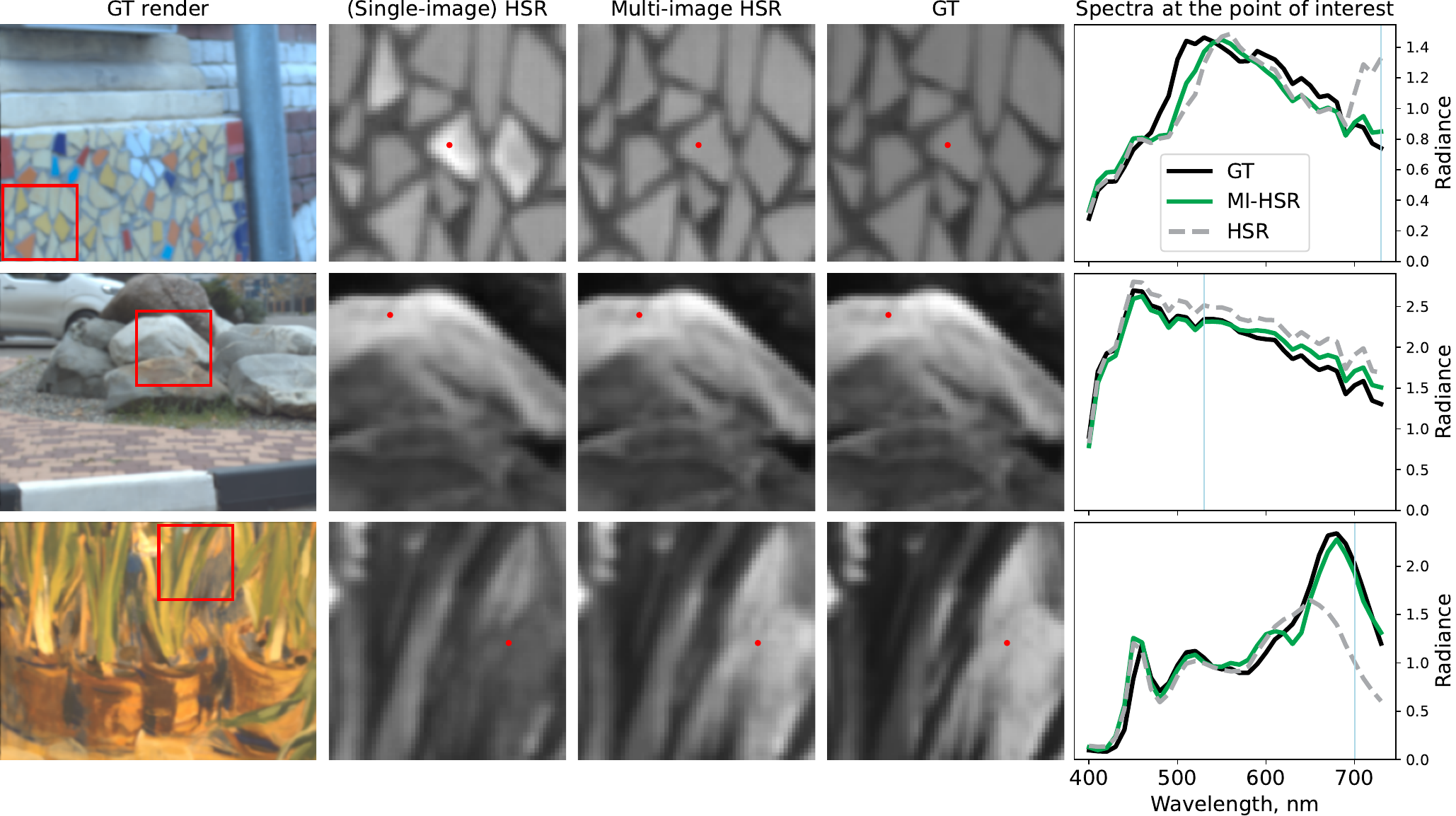}
    \caption{Qualitative comparison of GMST++ and DCAM + GMST++ predictions under single-image and multi-image input settings respectively. For each scene, a specific spectral band, region of interest, and point of interest are selected.
    Grayscale renderings correspond to a specific spectral band (indicated in blue on spectral plot). Colorful GT renderings were synthesized using the CIE 1931 RGB color matching functions.}
    \label{fig:qualitative_eval}
\end{figure*}

In the Real-world setting, we observed that training stochasticity introduces considerable variability in the final evaluation metrics, which we attribute to the relatively small size of our dataset.
To ensure a statistically robust comparison of the methods, we repeated each training procedure 10 times and report the mean and double standard deviation (\(\approx 95\%\) confidence interval) across these runs in Table~\ref{tab:real-world}.
For multi-image input our DCAM + GMST++ method reaches 34.47~dB~PSNR, which is +0.65~dB higher than its nearest competitor and at least +2.92~dB better than any method operating on single-image input (it is also worth to take into account irreducible error, caused by Specim IQ noise, see Sec.~\ref{sec:eval:noise}).
The latter improvement secures applicability of multi-image approach in practice.

We show qualitative performance of our model based on patch-level comparisons at a selected wavelength as well as radiance profiles at specific points of interest in Fig.~\ref{fig:qualitative_eval}.
The multi-camera system consistently recovers more plausible surface details than the single-camera setup.
In the first and third rows, single-camera pipeline misestimates a specific region at a specific wavelength band (730 and 700 nm respectively).
In the second row, single-camera pipeline lags behind in overall estimation of brightness, though correctly recovering the shape of the spectrum.
These results demonstrate that multi-view fusion improves both spatial and spectral fidelity in the MI-HSR system.

\subsection{Ablation Study}
\label{sec:eval:ablation}
\begin{wraptable}[7]{r}{0.4\linewidth}
\vspace{-4em}
\resizebox{\linewidth}{!}{
\begin{tabular}{@{}ccrrr@{}}
\toprule
DCAM & GDFN & PSNR, dB ↑ & SAM, ° ↓ & NSE, \% ↓ \\
\midrule
  &   & 33.19 ± 0.86 & 3.99 ± 0.12 & 9.66 ± 0.80 \\
  & + & 33.47 ± 0.59 & 4.03 ± 0.24 & 9.45 ± 0.63 \\
+ &   & 33.82 ± 0.30 & 3.95 ± 0.25 & 9.09 ± 0.44 \\
+ & + & \textbf{34.47 ± 0.77} & \textbf{3.85 ± 0.25} & \textbf{8.45 ± 0.82} \\
\bottomrule

\end{tabular}}
\caption{Impact of various architecture components on the overall reconstruction quality. Evaluated on the Doomer dataset.
\textbf{Best} results are highlighted.}
\label{tab:ablation-of}
\end{wraptable}
To assess importance of each component in our model, we perform an ablation study by either replacing GDFN with Feed-Forward Network (fallback to MST++) or removing the DCAM module along with the optical flow network in it.
All evaluations are performed under Real-world multi-input setting and are shown in Table~\ref{tab:ablation-of}.

\subsection{Adaptation to Different Smartphones}

So far, the network has been trained and evaluated on a specific triple-camera
system. To apply it to other phones without retraining, we condition it on the
cameras that produced its input. Each camera is described by a fingerprint of its
color response, following CCMNet~\citep{kim2025ccmnet}, and by its exposure level.

The model is trained only on synthetic inputs: hyperspectral images from Doomer and ARAD~1K~\citep{ntire-2022} rendered through the spectral sensitivities of 93 measured devices~\citep{smartphone-sens,gu-sens} and their mixtures, with the selected filters applied to two of the three cameras.
Training and test cameras are disjoint.
On these renders, moving from training to held-out cameras costs at most 0.57\,dB and 0.43$^\circ$, so performance is stable across simulated cameras.

We test on three phones fitted with the same two filters, with Specim~IQ ground truth:  27 test scenes of Doomer (Huawei Mate 40 Pro) and a small set of Huawei Mate~80~Pro and seven iPhone~16~Pro photos.
No real RGB image from these phones is used in training.
We compare the multi-image model with a single-image baseline that reads only the main camera.
Results are presented in Table~\ref{tab:physical-transfer}.
The model transfers to all three phones without retraining or fine-tuning, although each evaluation set is small.
On the Mate~40~Pro and the Mate~80~Pro, the multi-image model outperforms the single-image baseline on all three metrics, reducing SAM from 18.55$^\circ$ to 9.56$^\circ$ and from 12.96$^\circ$ to 5.33$^\circ$.

\begin{table}[ht]
\resizebox{\linewidth}{!}{
\centering
\begin{tabular}{@{}lrrrrrrrrr@{}}
\toprule
 & \multicolumn{3}{c}{Huawei Mate 40 Pro} & \multicolumn{3}{c}{Huawei Mate~80~Pro} & \multicolumn{3}{c}{iPhone~16~Pro} \\
\cmidrule(lr){2-4} \cmidrule(lr){5-7} \cmidrule(l){8-10}
Method & PSNR, dB ↑ & \multicolumn{1}{c}{SAM, ° ↓} & \multicolumn{1}{c}{NSE, \% ↓}
 & PSNR, dB ↑ & \multicolumn{1}{c}{SAM, ° ↓} & \multicolumn{1}{c}{NSE, \% ↓}
 & PSNR, dB ↑ & \multicolumn{1}{c}{SAM, ° ↓} & \multicolumn{1}{c}{NSE, \% ↓} \\
\midrule
Single-image & 20.02 & 18.55 & 32.31 & 11.86 & 12.96 & 25.05 & 17.81 & \textbf{5.86} & \textbf{18.97} \\
Multi-image & \textbf{26.33} & \textbf{9.56} & \textbf{17.64} & \textbf{20.38} & \textbf{5.33} & \textbf{13.69} & \textbf{23.23} & 9.12 & 20.29 \\
\bottomrule
\end{tabular}
}
\vspace{0.2em}
\caption{Transfer of models conditioned on the smartphone camera fingerprint trained only on synthetic data to three physical
phones. The better value per device is in bold.}
\label{tab:physical-transfer}
\end{table}
On the iPhone, the multi-image model has higher NSE and SAM than the single-image baseline (9.12$^\circ$ vs.\ 5.86$^\circ$), although its PSNR is higher.
Two factors explain this.
First, auto-exposure left the telephoto camera very dark, so its channels have a low signal-to-noise ratio.
Second, the network is not trained to identify which input channels are reliable, so it cannot down-weight such a channel.
A larger evaluation is required to confirm these trends.
\section{Limitations}
\label{sec:limitations}

Despite multi-image input improves reconstruction quality over single-image input both in the Clean (Table~\ref{tab:clean-world}) and Real-world (Table~\ref{tab:real-world}) settings, this gain comes at a substantial computational cost, primarily due to the necessity of optical flow.
DCAM + GMST++ requires about $3\times$ more MACs than the single-image GMST++.
The current model is therefore not suited to real-time inference on mobile devices.

In addition, Doomer contains only static scenes.
While some camera APIs support simultaneous capture from several cameras, in our setup the three smartphone images were captured sequentially, and some hyperspectral scans took tens of seconds.
The method's performance on dynamic scenes therefore remains unverified.

Finally, cross-device generalization was evaluated mainly on synthetic data and on a limited number of physical devices.
A dataset pairing hyperspectral ground truth with captures from \emph{many} different RGB sensors would allow a more thorough assessment.

\section{Conclusion}
\label{sec:conclusion}
This work rethinks HSR through the lens of practical acquisition.
By moving beyond single RGB image constraints, we demonstrate that leveraging multiple smartphone cameras with carefully chosen spectral filters can significantly enrich the input signal and reduce ambiguity in reconstruction.

The introduction of the Doomer dataset marks an important step forward for benchmarking in this space, enabling systematic evaluation of multi-view HSR under realistic conditions.
Empirical results validate both our hardware configuration and model design, suggesting a promising direction for low-cost, deployable hyperspectral imaging systems.
Proposed Deformable Convolution Alignment Module and conducted experiments show the alignment importance in the MI-HSR task, while proposed GMST++ is comparable with other state-of-the-art methods in established single image HSR it shows spectral reconstruction improvement in multi-image input even without DCAM.
Finally, easy-to-implement adaptation to other triple-camera smartphone was demonstrated.

Looking ahead, we aim to further explore the temporal dimension for dynamic scenes, optimize for energy-efficient mobile deployment, and investigate more principled learning paradigms under limited supervision or device mismatch.

\section*{Acknowledgments}
The work was supported by the Alexander von Humboldt Foundation and the Ministry of Economic Development of the Russian Federation (agreement No. 139-15-2025-013, dated June 20, 2025, IGK 000000C313925P4B0002).

{
    \small
    \bibliographystyle{unsrtnat}
    \bibliography{bibliography}
}

\newpage
\appendix
\section{Technical Appendices and Supplementary Material}
To ensure disambiguation, here is the list of some designations used throughout the main paper and this document:
\begin{itemize}
    \item Photometric normalization --- division of a camera sensor signal by ISO (if applicable) and exposure time. Black current subtraction is also applied beforehand.
    \item $\mathbf{x}$ --- a photometrically normalized signal (probably noisy) from smartphone camera(s). Can be $\in \mathbb{R}^3$ or $\in \mathbb{R}^{3k}$ depending on context.
    \item $\bar{\mathbf{x}}$ --- a photometrically normalized signal as it would be in an ideal world without noise.
    \item $\mathbf{y}$ --- a photometrically normalized radiance spectrum. Is assumed to be noiseless (however, in Sec.~\ref{sec:eval:noise} we discuss the outcomes of this assumption).
\end{itemize}
\subsection{Estimation of Spectral Sensitivity Functions and Transmittance Functions}
\label{sec:supp:measurements}

\begin{figure}[h]
    \centering
    \begin{subfigure}{.49\linewidth}
        \centering
        \includegraphics[width=\textwidth]{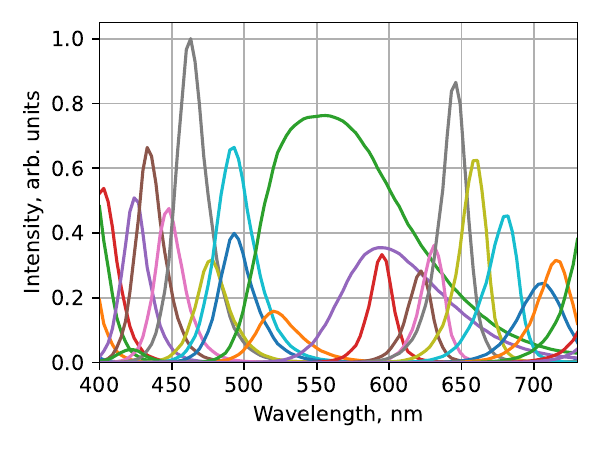}
        \caption{LEDs' radiance spectra}
    \end{subfigure}
    \begin{subfigure}{.49\linewidth}
        \centering
        \includegraphics[width=\textwidth]{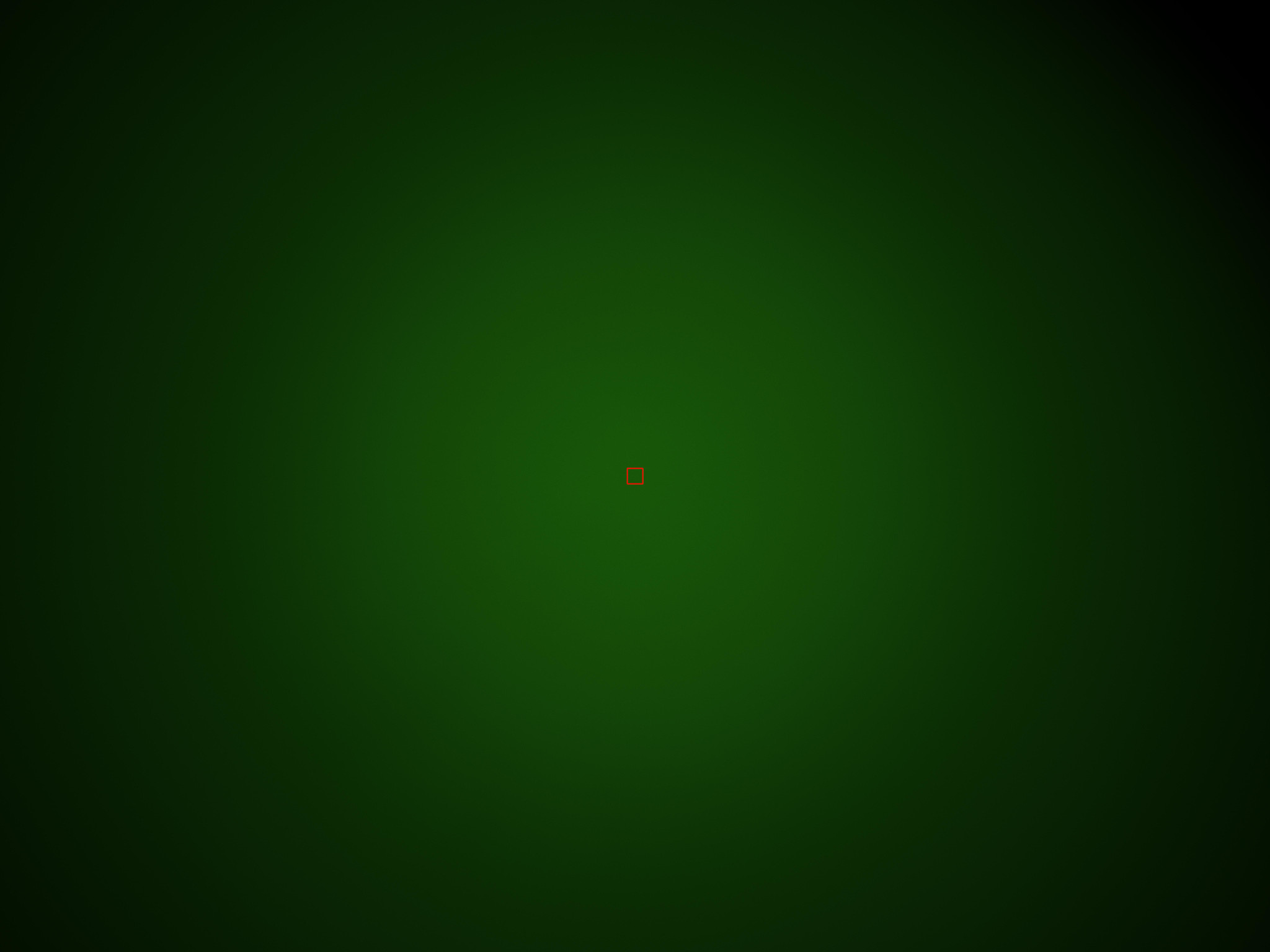}
        \caption{Sample LED FFI photo}
    \end{subfigure}%
    \caption{Training data for SSF estimation: (a) spectra of 25 LEDs measured using X-Rite i1 Pro; (b) an example of raw FFI photo. Notice the vignetting effect. To mitigate this effect, channel-wise average of a tiny central patch (in red) was used as a sample $\vx_i$}
    \label{fig:sens-samples}
\end{figure}

\paragraph{Spectral sensitivity functions}
To estimate SSF of a smartphone camera, we acquired 25 sample pairs of $\vx_i \in \mathbb{R}^3$ and $\vy_i \in \mathbb{R}^n$ corresponding to flat-field illumination (FFI) of narrow-band LED light sources in an integrating sphere.
For each LED $i$, we took a photo of FFI (Fig.~\ref{fig:sens-samples}), extracted the central 100×100 patch from it, photometrically normalized it and computed the channel-wise average to get $\vx_i$.
Then we measured the LED radiance spectrum $\vy_i$ using an X-Rite i1 Pro spectrophotometer with the help of \texttt{spotread} routine from Argyll color management system.

Given the measurements $\{\vx_i, \vy_i\}_{i=1}^{25}$, the spectral sensitivities estimation problem can be formulated as a regularized quadratic optimization problem:
\begin{align}
\label{eq:sens-est}
\begin{aligned}
    \min_\mC &\quad \sum_{i=1}^{25} \Vert \vx_i - \mC^\top\vy_i \Vert_2^2 +
    \lambda \Vert \mD\mC \Vert_2^2\\
    \text{s. t.} &\quad \mC \ge 0
\end{aligned}
\end{align}
where the regularization term $\lambda \Vert \mD\mC \Vert_2^2$ imposes smoothness, $\mD$ is the second-order derivative operator.
The objective (\ref{eq:sens-est}) was minimized using the Adam optimizer.

\begin{figure}[h]
    \centering
        \begin{subfigure}[b]{.33\textwidth}
        \centering
        \includegraphics[width=\textwidth]{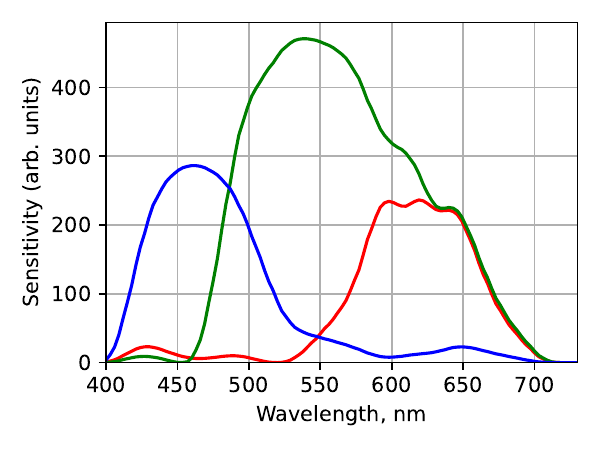}
        \vspace{-2em}
        \caption{Main camera sensitivities}
    \end{subfigure}%
    \begin{subfigure}[b]{.33\textwidth}
        \centering
        \includegraphics[width=\textwidth]{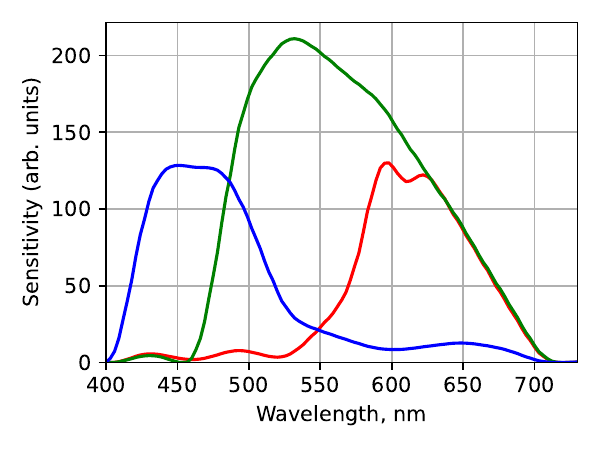}
        \vspace{-2em}
        \caption{Tele camera sensitivities}
    \end{subfigure}%
    \begin{subfigure}[b]{.33\textwidth}
        \centering
        \includegraphics[width=\textwidth]{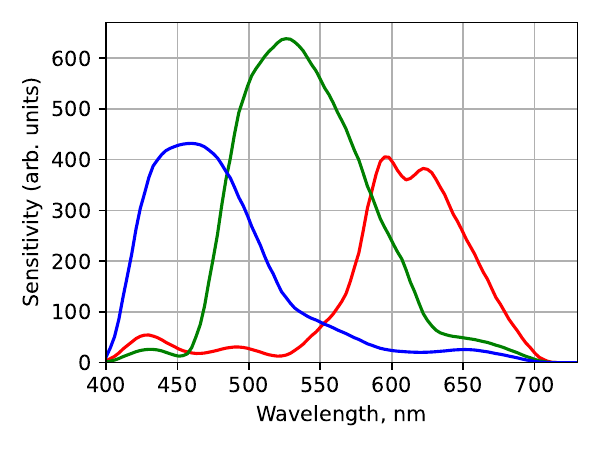}
        \vspace{-2em}
        \caption{Wide camera sensitivities}
    \end{subfigure}
    \label{fig:camera-ssens}
    \caption{Estimated smartphone cameras spectral sensitivities.}
\end{figure}

To validate the estimated $\hat{\mC}$, we captured a color rendition chart by the Specim IQ hyperspectral camera and all three smartphone cameras.
The HSI was calibrated (more on that in the next paragraph) and projected onto each camera's sensor space to yield RGB predictions.
We compared the predicted and actual RGBs of the color patches: pixels of a patch were pixel-wise averaged (Fig.~\ref{fig:sigma-mean} gives a sight on how such patches look like).
The average angular error of color reproduction was $\approx 1^\circ$.

\paragraph{Specim IQ calibration}
Spectral measurements of X-Rite i1 Pro and Specim IQ are actually inconsistent.
It is not surprising: Specim IQ is not designed to be photometrically accurate.
Instead, it allows to estimate band-wise relations between the \emph{calibration target} in a scene and other objects.
However, if we estimate the band-wise calibration divisor between X-Rite i1 Pro and Specim IQ, we can further use it to obtain physically-correct HSIs.
We illuminated the integration sphere using a mixture of LEDs that yields even spectral power distribution, obtained X-Rite and Specim IQ measurements of it and divided one by another:
\begin{equation*}
    \vk = \vy_{\mathrm{specim}}\oslash\vy_{\mathrm{xrite}},
\end{equation*}
The divisor is applied to every HSI filmed by the Specim IQ in this research.
\begin{figure}[h]
    \centering%
    \includegraphics[width=0.5\linewidth]{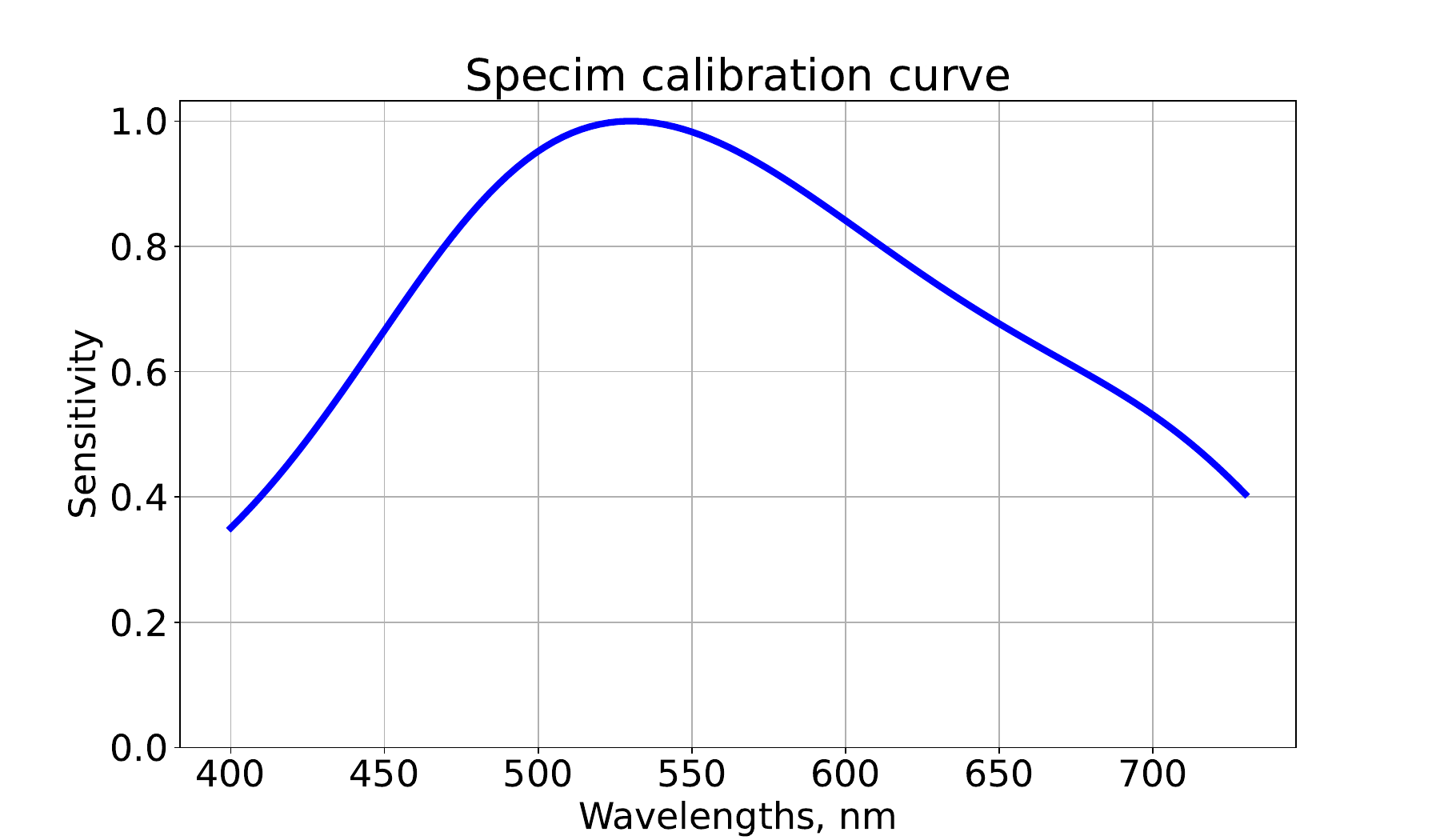}%
    \caption{The estimated Specim IQ calibration divisor}%
    \label{fig:specim_sens}%
\end{figure}%

Since the operational range of Specim IQ is 400--1000~nm and the one of X-Rite is 380--730~nm, all our spectral measurements are defined on their intersection: 400--730~nm.

\textbf{Transmittance functions of filters} were measured using an SF-2000 spectrophotometer.

\subsection{Spectral Uncertainty}
\label{sec:supp:filters}
The inspiration comes from the well-known conditional entropy $H\left(\xi \mid \eta\right)$, which quantifies the information loss of a latent random variable $\xi$ when observing the outcomes of another random variable $\eta$:
\begin{equation}
\label{eq:entropy-def}
H\left(\xi \mid \eta\right) = \mathbb{E}_{x\sim \eta}\, H\left(\xi \mid \eta = x\right),
\end{equation}
where $H\left(\xi \mid \eta = x\right)$ is the ordinary entropy of $\xi$ given event $\eta = x$.
However, entropy is designed especially for discrete-distributed variables of categorical type.
When uncertainty of a categorical random variable is indeed best described by the entropy of its distribution, for continuous random variables variance is a better choice.
When dealing with random vectors, variance becomes a matrix, so we consider the trace of it as a reasonable summary.
If we put the trace of variance instead of the entropy in (\ref{eq:entropy-def}) and substitute $\xi=\mathbf{y}, \eta=\mathbf{x}$, we get the spectral uncertainty of our filters-modified optical system:
\[
v(\mF) = \mathbb{E}_{\mathbf{x}} \left[ \mathrm{tr} \, \mathrm{Var}_{\mathbf{y}}(\mathbf{y} \mid \mathbf{x}) \right],
\]
where the dependence on $\mF$ is hidden inside the relationship between $\mathbf{y}$ and $\mathbf{x}$.
However, $\mathbf{y}$ and $\mathbf{x}$ are not random variables until we define them so.
Let $\mathbf{y}$ have a discrete distribution $p\left(\mathbf{y}_i\right)$ over a finite set of radiance spectra $\mathcal{R}=\{\mathbf{y}_1, ..., \mathbf{y}_N\}$ derived from a dataset of HSIs.
Such definition reflects the \emph{a priori} information about natural radiance.
We also model $\mC_\mF$ as a random convex combination of SSFs $\mC^{(1)}$, ..., $\mC^{(7)}$ under our filters: $\mC^{(1)}_\mF$, ..., $\mC^{(7)}_\mF$.
The coefficients are sampled from $\mathcal{U}[0, 1]$ and then re-normalized to sum up to 1.
In the relation $\mathbf{x}= \mC_\mF^\top \mathbf{y+n}$, only noise $\mathbf{n}$ is yet to define.
Let $\bar{\mathbf{x}} = \mC_\mF^\top\mathbf{y}$ be an unnoised camera response. 
We model noise as
\begin{gather}
\label{eq:noise-intro}
\mathbf{n}\mid\bar{\mathbf{x}} \; \sim \;
\mathcal{N}(\mathbf{0}, \mathbf{\Sigma(\bar{\mathbf{x}})}),\\
\label{eq:sigma-intro}
\mathbf{\Sigma}(\bar{\mathbf{x}})=\mathrm{diag}[\sigma^2_1(\bar{c}_1); ...; \sigma^2_{3k}(\bar{c}_{3k})],
\end{gather}
where $\sigma_i(\cdot)$ is expanded in (\ref{eq:sigma-def}) later in the chapter.

Now, when we defined $p\left(\mathbf{x}\mid\mathbf{y}\right)$, we can derive from the Bayesian rule:
\[
p\left(\mathbf{y}_i\mid\mathbf{x}\right) = \frac{
    p\left(\mathbf{x}\mid\mathbf{y}_i\right) p\left(\mathbf{y}_i\right)
}{
    \sum_{i=1}^N p\left(\mathbf{x}\mid\mathbf{y}_i\right) p\left(\mathbf{y}_i\right),
}
\]
and further:
\begin{gather*}
\label{eq:sum1}
\mathbb{E}_\mathbf{y}\left( \mathbf{y} \mid \mathbf{x} \right) =
\sum_{i=1}^N p\left(\mathbf{y}_i\mid\mathbf{x}\right) \mathbf{y}_i,
\\
\label{eq:sum2}
\mathrm{tr}\,\mathrm{Var}_\mathbf{y}\left( \mathbf{y} \mid \mathbf{x} \right) =
\sum_{i=1}^N p\left(\mathbf{y}_i\mid\mathbf{x}\right)
\left\Vert
    \mathbf{y}_i - \mathbb{E}_\mathbf{y}\left( \mathbf{y} \mid \mathbf{x} \right)
\right\Vert^2_2.
\end{gather*}

Now we have an expression depending on $\mathbf{x}$ which we should take $\mathbb{E}_\mathbf{x}$ of.
However, this is intractable.
So we should resort to Monte Carlo method by sampling $\mathbf{x}$ according to its definition: first take random $\mathbf{y}^* \in \mathcal{R}$, then sample $\mC_\mF$, then sample $\mathbf{x}$. In our experiments, $2^{20}$ samples were sufficient to achieve 1\% of relative standard deviation.

To derive $\mathcal{R}$, we first extracted every 29×29th pixel of the KAUST dataset~\citep{kaust-dataset}, which resulted in 559,921 samples.
To reduce computational requirements, we compressed $\mathcal{R}$ to the size of $N = 1024$ by running K-Means algorithm and employing clusters' centers.
We assigned $p\left(\mathbf{y}_i\right)$ to be the share of initial 559,921 radiance spectra that belong to the cluster around $\mathbf{y}_i$.
The values of $\mC^{(1)}$, ... $\mC^{(7)}$ are given by publicly available SSFs of Sony~ILCE-7, Samsung~NX1100, Lumix~GH2, Canon~6D and each all our three measured SSFs respectively.


\begin{figure}
    \centering
    \includegraphics[width=0.3\textwidth]{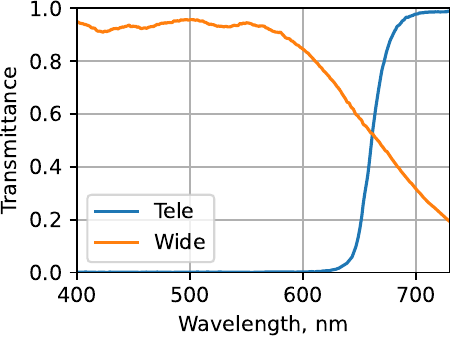}
    \caption{Filters selected for Tele and Wide cameras.}
    \label{fig:best-filters}
\end{figure}


\paragraph{Experimental justification} Having the estimated spectral characteristics of our filtered smartphone cameras setup ($\mC_\mF$) and the sensor noise model (\ref{eq:sigma-def}), we find ourselves in a position where we can simulate images of the triple-camera setup from HSIs and run experiments on simulated data.
Therefore, we can support the proposed spectral uncertainty criterion by simulating input images $\tX_0$, $\tX_1$, $\tX_2$ given filters, measuring HSR performance and plotting it against $v(\mC_\mF)$.

\begin{figure}[ht]
\centering
\begin{subfigure}[b]{0.33\linewidth}
    \centering
    \includegraphics[width=\textwidth]{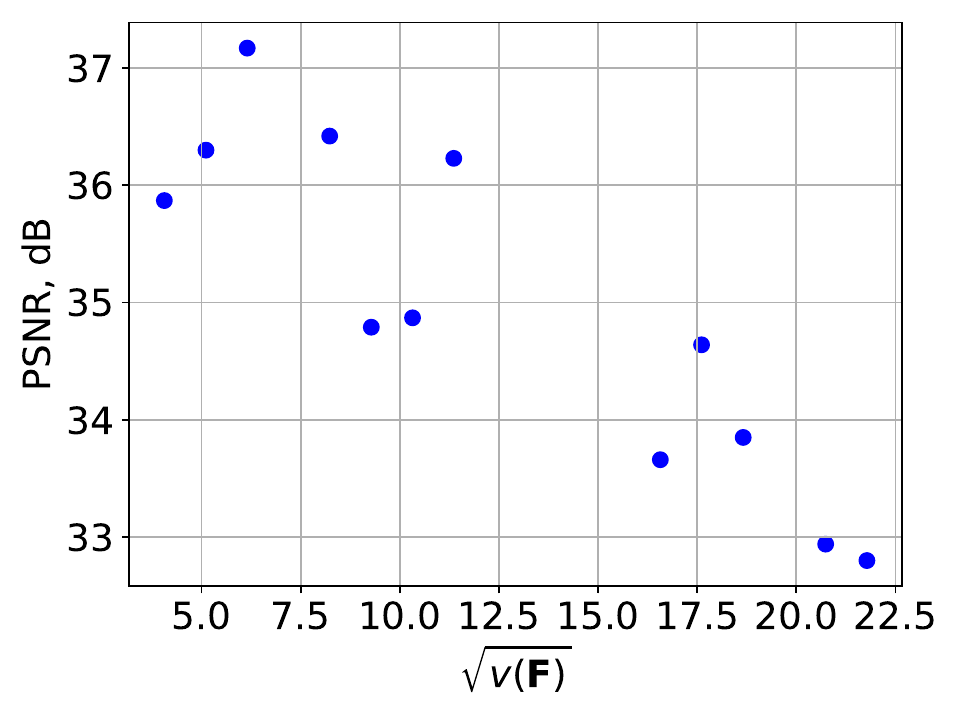}
    \vspace{-1.5em}
    \caption{PSNR vs Variance}
    \label{fig:psnr-var}
\end{subfigure}%
\begin{subfigure}[b]{0.33\linewidth}
    \centering
    \includegraphics[width=\textwidth]{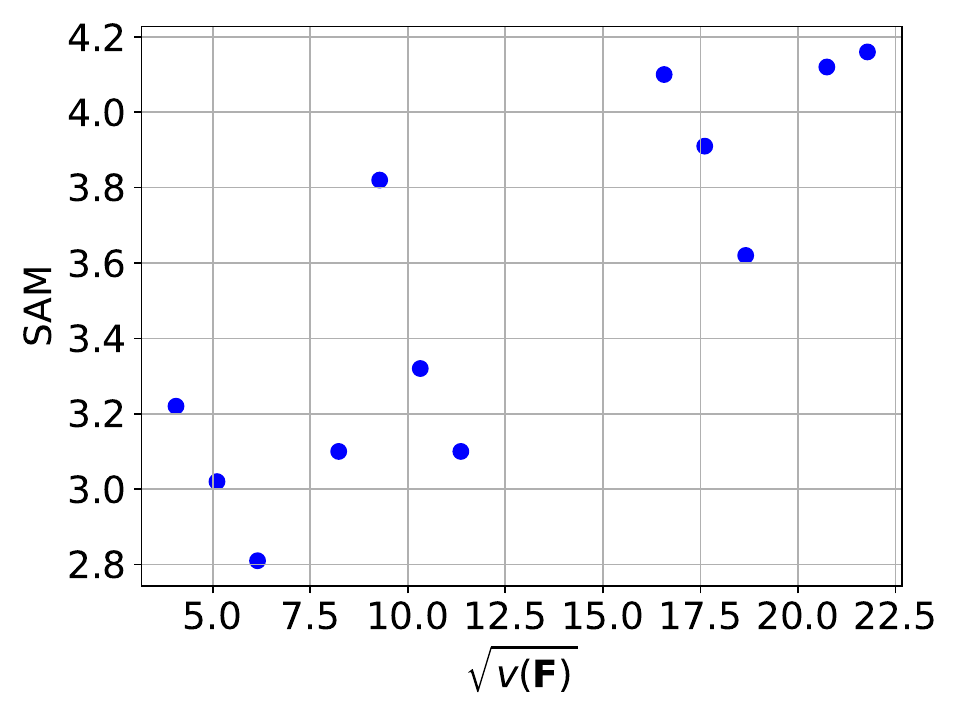} 
    \vspace{-1.5em}
    \caption{SAM vs Variance}
    \label{fig:sam-var}
\end{subfigure}%
\begin{subfigure}[b]{0.33\linewidth}
    \centering
    \includegraphics[width=\textwidth]{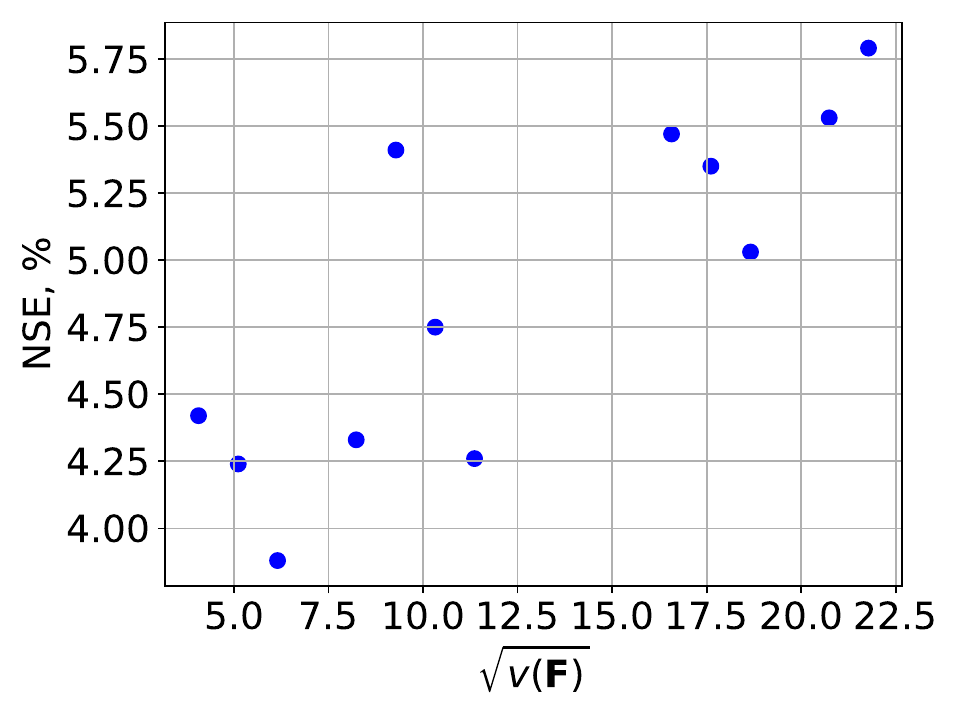} 
    \vspace{-1.5em}
    \caption{NSE vs Variance}
    \label{fig:nse-var}
\end{subfigure}
\caption{Correlation between PSNR, SAM, NSE (simulated settings) and the proposed spectral uncertainty.}
\label{fig:metrics-vs-variance}
\end{figure}

We chose 12 random filter pairs and simulated 12 versions of our dataset.
For each version, MST++ was trained and evaluated. Fig.~\ref{fig:metrics-vs-variance} shows the dependence of three performance metrics on the proposed spectral uncertainty criterion.
A strong correlation is observed, as the Pearson correlation coefficient has an absolute value greater than 0.8.

\paragraph{Noise model}
Let $\bar{x}_i$ be a photometrically-normalized signal level at channel $i$, $t$ be exposure time, $g$ be ISO.
The expected collected charge at image sensor is $\bar{x}_it$.
Following~\citet{noise-model}, we model noise of the collected charge as $\varepsilon \sim \mathcal{N}(0, \alpha_i \bar{x}_i t + \beta_i)$, where $\alpha_i$, $\beta_i$ are parameters of the model.
The final value in the raw image is further amplified by $g$ along with the noise: $(\bar{x}_i t + \varepsilon)g$.
Foi et al. also introduce additional noise after amplification, but we neglect it for simplicity.
The observed photometrically-normalized signal is then given by:

\[
x_i = \frac{(\bar{x}_i t + \varepsilon)g}{tg} = \bar{x}_i + \frac{\varepsilon}{t}.
\]

To be consistent with (\ref{eq:noise-intro}) and (\ref{eq:sigma-intro}), we define $\sigma_i(x)$ as:
\begin{equation}
\label{eq:sigma-def}
\sigma_i(\bar{x}_i) = \frac{\sqrt{\alpha_i \bar{x}_i t + \beta_i}}{t}
\end{equation}

\begin{figure}[h]
    \centering
    \includegraphics[align=c,width=0.30\linewidth]{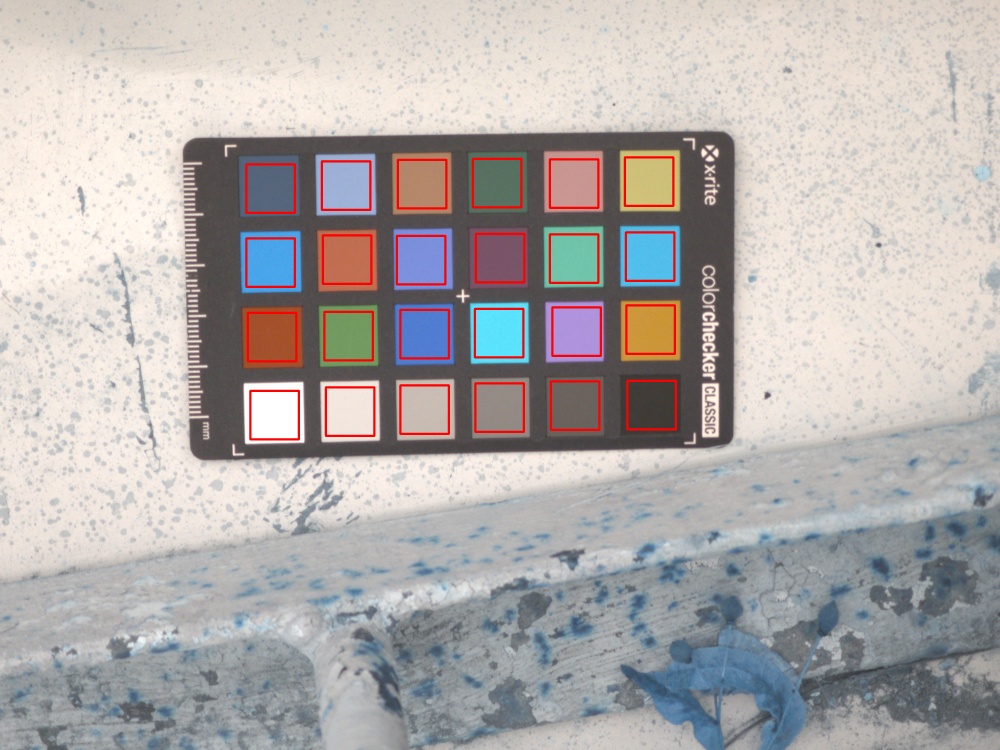}%
    \includegraphics[align=c,width=0.70\linewidth]{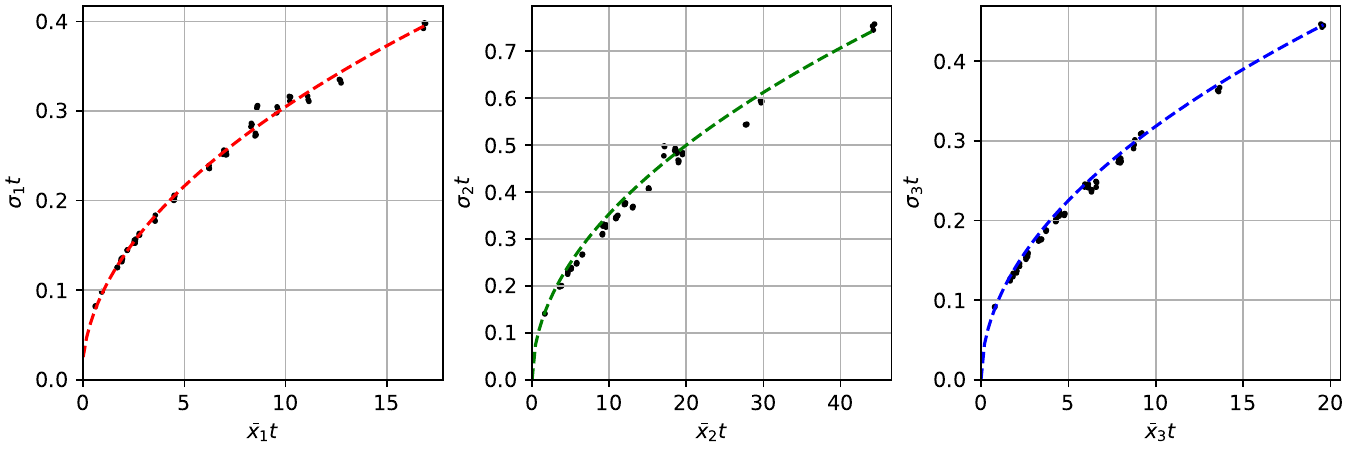}
    \caption{
Plotting $\sigma_i t$ against $\bar{x}_i t$ for the Telephoto camera based on a single shot.
Each data point (24 in total) is given by channel-wise mean and std dev of a color patch (left, in red).
In each plot, point cloud is approximated by a dashed line (\ref{eq:sigma-def}) using least-squares; each color channel $i$ gets its own noise parameters $\alpha_i$, $\beta_i$ (right).
    }
    \label{fig:sigma-mean}
\end{figure}

We estimate parameters $\alpha_i$, $\beta_i$ of this model by plotting $\sigma_i t$ against $\bar{x}_i t$ in Fig.~\ref{fig:sigma-mean}.

\begin{wrapfigure}[12]{r}{0.3\textwidth}
    \vspace{-2em}
    \centering
    \includegraphics[width=\linewidth]{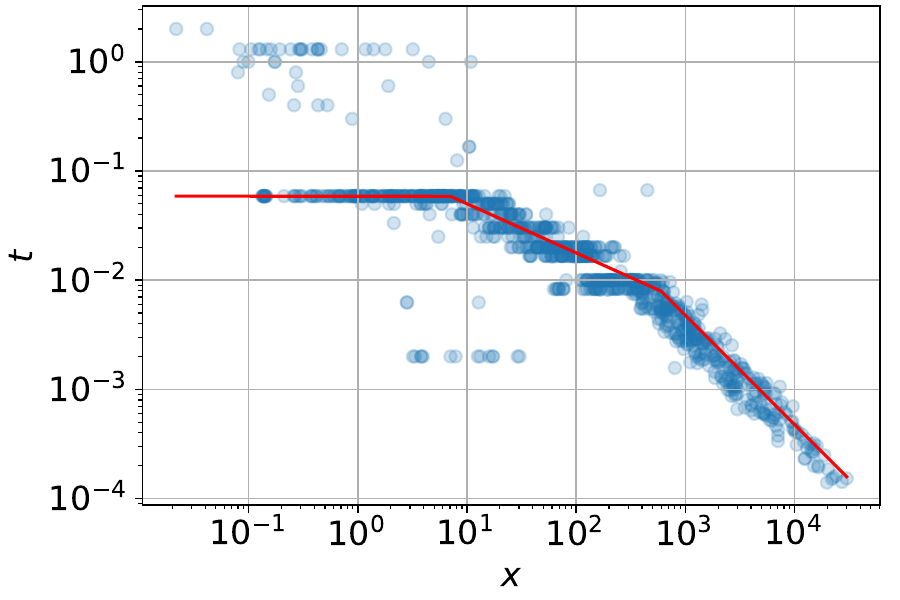}
    \caption{How exposure time depends on scene brightness. The piecewise-power function plotted in red line is used to model the exposure time given the average of a photometrically normalized image.}
    \label{fig:exposure-model}
\end{wrapfigure}

The only problem left in (\ref{eq:sigma-def}) is the hidden dependence on $t$.
Given simulated $\bar{\mathbf{x}} = \mC_\mF^\top\mathbf{y}$, we cannot add noise properly unless we know exposure time $t$.
In real-life setting, exposure time is determined by a camera automatically depending on the brightness of a scene.
For simplicity, we model this in a piecewise-power fashion.
We gathered 1000 raw images from phone's gallery and plotted $t$ against the average of photometrically normalized signals across all channels, height and width in an image $x = \frac{1}{h\times w\times 3}\sum_{i, j, k} \tX_{i, j, k}$ (Fig.~\ref{fig:exposure-model}).
The fitted function allows to estimate $t$ given $\bar{x}_i$.

\subsection{Related Datasets}
\label{sec:datasets}
Table~\ref{tab:datasets} shows an overview of existing datasets for HSR from RGB. Our proposed Doomer dataset uniquely offers real multi-view RGB images with spectral filters, misalignment, and in-scene gray reference under diverse conditions.

\begin{table}[!ht]
\centering
\resizebox{1.0\linewidth}{!}{
\begin{tabular}{@{}l r c c c c @{}}
\toprule
\textbf{Dataset} & \textbf{\begin{tabular}[c]{@{}c@{}}\# scenes\end{tabular}} & \textbf{Spectral data} & \textbf{Color reference} & \textbf{\begin{tabular}[c]{@{}c@{}}Spectral sampling\\ Range (step), nm\end{tabular}} & \textbf{\begin{tabular}[c]{@{}c@{}}Corresponding\\RGBs\end{tabular}} \\ \midrule
CAVE \citep{cave-dataset} & 32 & Reflectance & Color chart or no & \begin{tabular}[c]{@{}c@{}}400--700 (10)\end{tabular} & simulated BMP \\
Harvard \citep{harvard-dataset} & 79 & Radiance & No & \begin{tabular}[c]{@{}c@{}}400--720 (10)\end{tabular} & No \\
ICVL \citep{arad-icvl}& 200 & Radiance & Color chart or no & \begin{tabular}[c]{@{}c@{}}400--1000 (10)\end{tabular} & simulated JPEG \\
KAUST \citep{kaust-dataset}& 409 & Reflectance & White patch & \begin{tabular}[c]{@{}c@{}}400--730 (10)\end{tabular} & No \\
ARAD~1K \citep{ntire-2022}& 949 & Radiance & No & \begin{tabular}[c]{@{}c@{}}400--700 (10)\end{tabular} & simulated JPEG \\
Beyond RGB \citep{beyond-rgb}& 1680 & Radiance & Color chart & \begin{tabular}[c]{@{}c@{}}380--730 (20)\end{tabular} & 2× real RAW \\
MobileSpec \citep{mobile-spec} & 200 & Radiance & No & \begin{tabular}[c]{@{}c@{}}400--1000 (10)\end{tabular} & real RAW \\
\midrule
Doomer & 155 & Radiance & Gray ball & \begin{tabular}[c]{@{}c@{}}400--730 (10)\end{tabular} & 3× real RAW \\
\bottomrule
\end{tabular}
}
\vspace{0.2em}
\caption{
Overview of existing datasets for HSR from RGB.  Our proposed Doomer dataset uniquely offers real multi-view RGB images with spectral filters, misalignment, and in-scene gray reference under diverse conditions.}
\label{tab:datasets}
\end{table}

\subsection{More on Doomer Dataset}
\label{sec:supp:dataset}
In the full public release of Doomer dataset we will provide 3 main versions:
\begin{enumerate}
    \item \textbf{RAW version}. Images from all cameras before any preprocessing.
    \item \textbf{Single-camera version}.
    Pairs of a Main-camera photo and an HSI. The images are preprocessed and share the same FoV and spatial resolution.
    \item \textbf{Multi-camera version}.
    Quadruples of Main-, Tele- and Wide-camera photos and HSI. The images are preprocessed and share the same FoV and spatial resolution.
\end{enumerate}
Both the single- and multi-camera versions will also have two subversions of hyperspectral sampling interval: 3~nm or 10~nm.
A short summary of each version provided in Tab.~\ref{tab:dataset-versions}.

The preprocessing pipeline of RAW images consists of these steps: demosaicing, black current subtraction, flat field calibration and photometric normalization.
For hyperspectral images, preprocessing pipeline includes: black current subtraction, flat field calibration, radial distortion correction and photometric normalization.
The last step involves only division by the exposure time (Specim IQ has no ISO setting) and band-wise division by the calibration divisor $\vk$.

\begin{table}[ht]
\resizebox{1.0\linewidth}{!}{
\begin{tabular}{@{}ccccc@{}}
\toprule
\textbf{\begin{tabular}[c]{@{}c@{}}Dataset\\ version\end{tabular}} & \textbf{\begin{tabular}[c]{@{}c@{}}Smartphone\\ images\end{tabular}} & \multicolumn{2}{c}{\textbf{\begin{tabular}[c]{@{}c@{}}HSIs\\ spectral range (step)\end{tabular}}} & \textbf{FoV matching} \\ \midrule
RAW & Main, Tele, Wide & \multicolumn{2}{c}{\begin{tabular}[c]{@{}c@{}}Original 400--1000 nm (3 nm)\end{tabular}} & -- \\ \midrule
1-camera & \begin{tabular}[c]{@{}c@{}}Preprocessed Main\\ camera image\end{tabular} & \begin{tabular}[c]{@{}c@{}}Preprocessed\\ 400--730 nm (10 nm)\end{tabular} & \begin{tabular}[c]{@{}c@{}}Preprocessed\\ 400--730 nm (3 nm)\end{tabular} & \checkmark \\ \midrule
3-camera & \begin{tabular}[c]{@{}c@{}}\emph{Preprocessed and concatenated} \\ Main, Tele, Wide images\end{tabular} & \begin{tabular}[c]{@{}c@{}}Preprocessed\\ \emph{400--730 nm (10 nm)}\end{tabular} & \begin{tabular}[c]{@{}c@{}}Preprocessed\\ 400--730 nm (3 nm)\end{tabular} & \checkmark \\ \bottomrule
\end{tabular}
}
\vspace{0.2em}
\caption{Doomer Dataset versions}
\label{tab:dataset-versions}
\end{table}

For all HSIs, we will provide a manually annotated binary mask that specifies the location of the gray ball.
The reflectance spectrum of the gray ball, measured with the X-Rite spectrophotometer, will be included in the dataset.
Also, we will expand it by adding more scenes and annotate each scene with tags depending on its contents.

Unlike previously published datasets, Doomer contains multiple real RGB images along with ground-truth HSI and illumination reference.
This combination enables the exploration of various computational photography problems, such as:
\begin{enumerate}
    \item \textit{White point estimation} both in spectral and RGB forms.
    \item \textit{Illumination distribution estimation}.
    The need to estimate a distribution of illumination in a scene arises from the complexity of natural scenes.
    In such environments, a single global white point may be insufficient for accurate image processing and color correction \citep{illum-distr}.
    \item \textit{Color space transform}.
    Since there is a known white-point for each scene, it is possible to do precise chromatic adaptation and color signals for different cameras.
    \item \textit{Hyperspectral Reconstruction}.
    Reconstruction from single or multiple cameras.
    Reconstruction of different types of spectra --- radiance or reflectance. 
\end{enumerate}

\begin{figure}[ht]
    \centering
    \includegraphics[width=\linewidth]{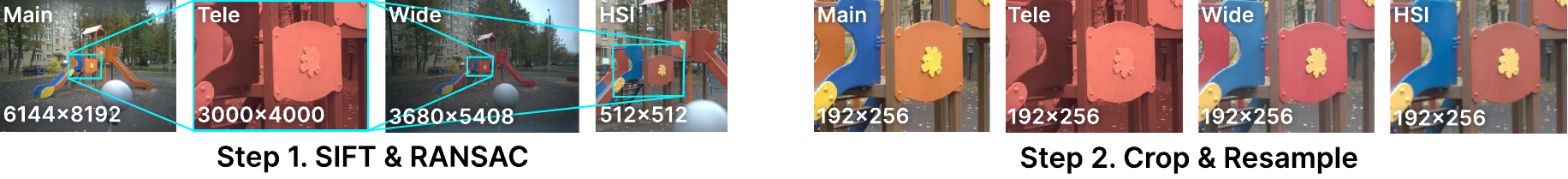}
    \caption{\textit{Spatial preprocessing pipeline}. Left: geometric alignment of RGB views using SIFT + RANSAC for consistent cross-camera registration. Right: field-of-view normalization and resolution matching across RGB and hyperspectral modalities.}
    \label{fig:preprocessing}
\end{figure}

The data preprocessing pipeline was described in Sec.~\ref{sec:dataset} and is briefly illustrated in Fig.~\ref{fig:preprocessing}.

\subsection{What is the Best Possible Quality?}
\label{sec:eval:noise}
Even professional hyperspectral equipment has its own precision, which limits best achievable quality as a GT.
Hyperspectral image registration is a process of registering photons, which means our GTs are inherently noisy just like consumer CMOS sensors.
To quantify this boundary, we captured series of HSIs of static scenes with color charts under two different light conditions.
The metrics were calculated between a fixed image from a series and the average of remain images, see Tab.~\ref{tab:specim-noise}.
The estimated values should be interpreted as approximate bound for any MI-HSR network trained and tested on Doomer dataset.
Comparable metric values may indicate overfitting or sensitivity to sensor noise.

\begin{table}[h]
\centering
\begin{tabular}{lccc}
\toprule
Light conditions & PSNR, dB ↑ & SAM, ° ↓ & NSE, \% ↓ \\
\midrule
Bright scenario (outdoor sunny) & 41.60 ± 3.41 & 0.71 ± 0.11 & 3.00 ± 0.71 \\
\midrule
Medium brightness (indoor LED) & 37.82 ± 3.51 & 1.02 ± 0.13 & 4.45 ± 1.28 \\
\bottomrule
\end{tabular}
\vspace{0.2em}
\caption{Mean and std of Specim IQ shots of the same scene.}
\label{tab:specim-noise}
\vspace{-1em}
\end{table}

\subsection{Failure Cases}

\begin{wrapfigure}[8]{r}{0.3\textwidth}
\vspace{-4em}
\includegraphics[width=\linewidth]{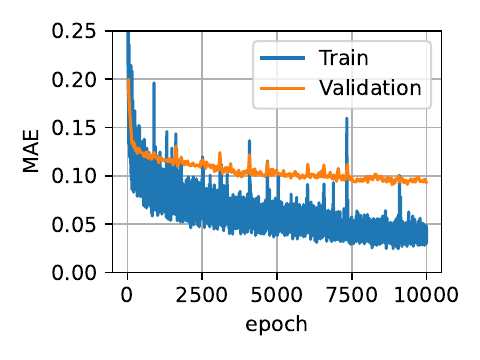}
\caption{Learning curves of our model on the Doomer dataset}
\label{fig:learning_curve}
\end{wrapfigure}

\begin{figure}
    \centering
    \begin{subfigure}[b]{.26\textwidth}
        \includegraphics[width=\textwidth]{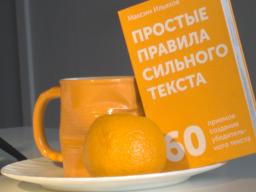}
        \caption{Main image}
    \end{subfigure}
    \begin{subfigure}[b]{.26\textwidth}
        \includegraphics[width=\textwidth]{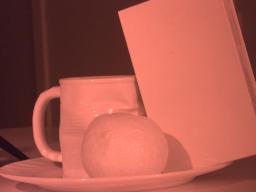}
        \caption{Telephoto image}
    \end{subfigure}
    \begin{subfigure}[b]{.26\textwidth}
        \includegraphics[width=\textwidth]{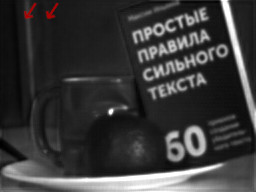}
        \caption{HSI prediction, \(\lambda= 410\) nm}
    \end{subfigure}
    \caption{Failure case of DCAM + GMST++ in the Real-world setting. The reconstructed HSI has a band, where the same background object is duplicated because of its displacement between input views.}
    \label{fig:failure-case}
\end{figure}

In cases of large object displacements between input views, significant occlusions or disocclusions may occur.
When a point is visible in one phone view but not in another, no valid correspondence exists, which can lead to ghosting artifacts in DCAM.
Figure~\ref{fig:failure-case} illustrates such a failure case, where a structure in the background is moving.
As a result, both instances of the same object remain visible in the reconstructed prediction.
A possible way to address this issue would be to introduce validity masks for the auxiliary images and use them to explicitly guide the reconstruction process.

\subsection{Does it Overfit?}

The training setup in the Real-world setting, involving 10,000 epochs and a dataset of only 155 scenes, raises concerns about potential overfitting.
First, we would like to clarify what we mean by an epoch here.
Our pipeline does not process every pixel in the dataset during a single epoch.
Instead, it processes only a random 64×64 patch from each of the training scenes.
Thus, in order to see the whole training dataset, at least \(\frac{256\times192}{64\times64}=12\) epochs are needed.
Second, we present an example of the learning curves for DCAM + GMST++ (Fig.~\ref{fig:learning_curve}).
This particular training run achieves a PSNR of 34.61~dB, exceeding the reported average of 34.47~dB, while showing no apparent signs of overfitting.
\subsection{Comparison of Optical Flow Models}
\label{sec:supp:of-ablation}

We evaluate the effect of switching between optical flow models inside DCAM in the Real-world setting.
We compare PWC-Net and RAFT-Small~\citep{raft}, justifying the ``-Small'' variant by having future mobile deployment in mind.
However, in our Doomer dataset every ground truth HSI is warped toward the reference image using PWC-Net, which may favor the PWC-Net-based DCAM.
Thus, for fair comparison we must cross-check and report the results under every variant of ground truth.
Since masks of valid ground truth pixels also depend on the optical flow model, they must be replaced with their intersection across the models.
Table~\ref{tab:supp-alignment} summarizes this experiment.
Although switching to a different optical flow model may improve certain metrics, this improvement is not systematic, and the confidence intervals overlap substantially.

\begin{table}[!ht]
\centering
\begin{tabular}{ccccc}
\toprule
\begin{tabular}[c]{@{}c@{}}Inside DCAM\end{tabular} & \begin{tabular}[c]{@{}c@{}}GT warping\end{tabular} & PSNR, dB ↑ & SAM, ° ↓ & NSE, \% ↓ \\
\midrule
PWC-Net & PWC-Net & \textbf{34.39 ± 0.77} & 3.86 ± 0.24 & 8.55 ± 0.71 \\
PWC-Net & RAFT-Small & 34.14 ± 0.75 & \textbf{3.85 ± 0.21} & 8.73 ± 0.82 \\
RAFT-Small & PWC-Net & 34.22 ± 0.60 & 3.96 ± 0.41 & 8.51 ± 0.57 \\
RAFT-Small & RAFT-Small & 34.34 ± 0.53 & 3.97 ± 0.23 & \textbf{8.43 ± 0.53} \\
\bottomrule
\end{tabular}
\vspace{0.2em}
\caption{Effect of switching between optical flow models inside DCAM and for GT warping.
Evaluated on the Doomer dataset in the Real-world setting.
\textbf{Best} results are highlighted.}
\label{tab:supp-alignment}
\end{table}

\subsection{Computational Resources}
\label{supp:compute}
All experiments were conducted on a cluster, which assigned one worker node with 16 CPU cores and one GPU per one training run.
The GPU could be one of the following: Nvidia A100, Nvidia L40 or Nvidia L40s.
Training of the proposed DCAM + GMST++ on Doomer dataset takes up to 2 hours.

\clearpage
\section*{NeurIPS Paper Checklist}

\begin{enumerate}

\item {\bf Claims}
    \item[] Question: Do the main claims made in the abstract and introduction accurately reflect the paper's contributions and scope?
    \item[] Answer: \answerYes{} 
    \item[] Justification: The abstract and introduction accurately describe the proposed smartphonebased spectral acquisition system, dataset (Doomer) and a neural network architecture for the MI-HSR task.

\item {\bf Limitations}
    \item[] Question: Does the paper discuss the limitations of the work performed by the authors?
    \item[] Answer: \answerYes{}
    \item[] Justification: The limitations described in the Sec.\ref{sec:limitations},\ref{sec:conclusion}, noting that current solution is hardly applicable for energy-efficient mobile deployment.

\item {\bf Theory assumptions and proofs}
    \item[] Question: For each theoretical result, does the paper provide the full set of assumptions and a complete (and correct) proof?
    \item[] Answer: \answerYes{} 
    \item[] Justification: We theoretically support our selection of filters briefly in Sec.~\ref{sec:filter-selection} and in detail in Supp.

    \item {\bf Experimental result reproducibility}
    \item[] Question: Does the paper fully disclose all the information needed to reproduce the main experimental results of the paper to the extent that it affects the main claims and/or conclusions of the paper (regardless of whether the code and data are provided or not)?
    \item[] Answer: \answerYes{} 
    \item[] Justification: The training details are partially given in Sec.~\ref{sec:exps-setup}. 

\item {\bf Open access to data and code}
    \item[] Question: Does the paper provide open access to the data and code, with sufficient instructions to faithfully reproduce the main experimental results, as described in supplemental material?
    \item[] Answer: \answerYes{} 
    \item[] Justification: The data and code will be published upon the paper acceptance.

\item {\bf Experimental setting/details}
    \item[] Question: Does the paper specify all the training and test details (e.g., data splits, hyperparameters, how they were chosen, type of optimizer) necessary to understand the results?
    \item[] Answer: \answerYes{} 
    \item[] Justification: We specify and briefly motivate the optimizer type and hyperparameters in Sec.~\ref{sec:exps-setup}. Data split is included in the code.

\item {\bf Experiment statistical significance}
    \item[] Question: Does the paper report error bars suitably and correctly defined or other appropriate information about the statistical significance of the experiments?
    \item[] Answer: \answerYes{} 
    \item[] Justification: The main results in the Table~\ref{tab:real-world} are averaged over 10 runs and report 2-sigma error bars.

\item {\bf Experiments compute resources}
    \item[] Question: For each experiment, does the paper provide sufficient information on the computer resources (type of compute workers, memory, time of execution) needed to reproduce the experiments?
    \item[] Answer: \answerYes{} 
    \item[] Justification: Specification of the required compute resources and execution time is given in Sec.~\ref{supp:compute}.
    
\item {\bf Code of ethics}
    \item[] Question: Does the research conducted in the paper conform, in every respect, with the NeurIPS Code of Ethics \url{https://neurips.cc/public/EthicsGuidelines}?
    \item[] Answer: \answerYes{} 
    \item[] Justification: The authors have fully complied the NeurIPS Code of Ethics

\item {\bf Broader impacts}
    \item[] Question: Does the paper discuss both potential positive societal impacts and negative societal impacts of the work performed?
    \item[] Answer: \answerNA{} 
    \item[] Justification: To date, there are no known social impacts imposed by hyperspectral reconstruction.
    
\item {\bf Safeguards}
    \item[] Question: Does the paper describe safeguards that have been put in place for responsible release of data or models that have a high risk for misuse (e.g., pre-trained language models, image generators, or scraped datasets)?
    \item[] Answer: \answerNA{} 
    \item[] Justification: To date, there are no known examples of hyperspectral data misuse.

\item {\bf Licenses for existing assets}
    \item[] Question: Are the creators or original owners of assets (e.g., code, data, models), used in the paper, properly credited and are the license and terms of use explicitly mentioned and properly respected?
    \item[] Answer: \answerYes{} 
    \item[] Justification: The paper properly cites the KAUST, ARAD 1K and spectral sensitivities datasets (licensed under CC BY-NC 3.0) used for filters selection and synthetic experiments.

\item {\bf New assets}
    \item[] Question: Are new assets introduced in the paper well documented and is the documentation provided alongside the assets?
    \item[] Answer: \answerYes{} 
    \item[] Justification: The assets will be best documented in the README file of the code.

\item {\bf Crowdsourcing and research with human subjects}
    \item[] Question: For crowdsourcing experiments and research with human subjects, does the paper include the full text of instructions given to participants and screenshots, if applicable, as well as details about compensation (if any)? 
    \item[] Answer: \answerNA{} 
    \item[] Justification: The paper does not involve crowdsourcing nor research with human subjects.

\item {\bf Institutional review board (IRB) approvals or equivalent for research with human subjects}
    \item[] Question: Does the paper describe potential risks incurred by study participants, whether such risks were disclosed to the subjects, and whether Institutional Review Board (IRB) approvals (or an equivalent approval/review based on the requirements of your country or institution) were obtained?
    \item[] Answer: \answerNA{} 
    \item[] Justification: The paper does not involve crowdsourcing nor research with human subjects.

\item {\bf Declaration of LLM usage}
    \item[] Question: Does the paper describe the usage of LLMs if it is an important, original, or non-standard component of the core methods in this research? Note that if the LLM is used only for writing, editing, or formatting purposes and does \emph{not} impact the core methodology, scientific rigor, or originality of the research, declaration is not required.
    \item[] Answer: \answerNA{} 
    \item[] Justification: No LLMs impacted the core methodology, scientific rigorousness, or originality of the research.

\end{enumerate}

\end{document}